\documentclass[12pt, hidelinks]{article}

\usepackage{amsmath, amssymb, url, float, graphicx, float,booktabs}
\usepackage{algorithm, algpseudocode}
\usepackage{fullpage}
\usepackage[small,bf]{caption}

\newcommand{\ones}{\mathbf 1}
\newcommand{\reals}{{\mbox{\bf R}}}

\newcommand{\symm}{{\mbox{\bf S}}}  %
\newcommand{\chol}{{\mbox{\bf chol}}}
\newcommand{\clip}{{\mbox{\bf clip}}}
\newcommand{\quantile}{{\mbox{\bf quantile}}}

\newcommand{\Tr}{\mathop{\bf Tr}}
\newcommand{\diag}{\mathop{\bf diag}}

\newcommand{\Expect}{\mathop{\bf E{}}}

\newcommand{\eg}{{\it e.g.}}
\newcommand{\ie}{{\it i.e.}}

\newcommand{\BEAS}{\begin{eqnarray*}}
\newcommand{\EEAS}{\end{eqnarray*}}
\newcommand{\BEA}{\begin{eqnarray}}
\newcommand{\EEA}{\end{eqnarray}}
\newcommand{\BEQ}{\begin{equation}}
\newcommand{\EEQ}{\end{equation}}
\newcommand{\BIT}{\begin{itemize}}
\newcommand{\EIT}{\end{itemize}}

\title{Covariance Prediction via Convex Optimization}
\author{Shane Barratt \and Stephen Boyd}
\date{\today}

\begin{document} 
\maketitle 
\begin{abstract}
We consider the problem of predicting the covariance of a 
zero mean Gaussian vector, based on another feature vector. 
We describe a covariance predictor that has
the form of a generalized linear model, \ie, an affine function of the features followed by an inverse link function 
that maps vectors to symmetric positive definite
matrices.  
The log-likelihood is a concave function of the predictor parameters,
so fitting the predictor involves convex optimization.
Such predictors can be combined with others, or 
recursively applied to improve performance.
\end{abstract}

\section{Introduction}
\label{s-introduction}

\subsection{Covariance prediction}\label{s-cov-pred}

We consider data consisting of a pair of vectors,
an outcome vector $y\in\reals^n$ and 
a feature vector $x\in \mathcal X \subseteq \reals^p$,
where $\mathcal X$ is the feature set.
We model $y$ conditioned on $x$ as
zero mean Gaussian, \ie, $ y \mid x \sim\mathcal N(0, \hat \Sigma(x))$,
where $\hat \Sigma : \mathcal X \to \symm_{++}^n$, the 
set of symmetric positive definite $n\times n$ matrices.
(We will address later the extension to nonzero mean.)
Our goal is to fit the covariance predictor $\hat \Sigma$ 
based on observed training data $x_i, y_i$, $i=1,\ldots,N$.
We judge a predictor $\hat \Sigma$ by its average log-likelihood on
out of sample or test data,
\[
\frac{1}{2 \tilde N} \sum_{i=1}^{\tilde N}  \left( 
-n \log (2\pi) - \log \det \hat \Sigma (\tilde x_i) - \tilde y_i^T 
\hat \Sigma (\tilde x_i) ^{-1}
\tilde y_i \right),
\]
where $\tilde x_i, \tilde y_i$, $i=1, \ldots, \tilde N$ is a test data set.

The covariance prediction problem arises in several contexts.
As a general example, let $y$ denote the prediction error 
of some vector quantity, using some prediction that depends on 
the feature vector $x$.  In this case we are predicting the 
(presumed zero mean Gaussian) distribution of the prediction error as a
function of the features.
In statistics, this is referred to as heteroscedasticity, since the uncertainty in the prediction depends on the feature vector $x$.

The covariance prediction problem also comes up in 
vector time series, where $i$ denotes time period.
In these applications, the feature vector $x_i$ is known in
period $i$, but the outcome $y_i$ is not, and we are predicting 
the (presumed zero mean Gaussian) distribution of $y_i$.
In the context of time series, the covariance predictor
$\hat \Sigma(x_i)$ is also known as the covariance forecast.
In time series applications,
the feature vector $x_i$ can contain quantities known at time $i$
that are related to $y_i$, including quantities that 
are derived directly from past values
$y_{i-1}, y_{i-2}, \ldots$, such as the entries of
$y_{i-1}y_{i-1}^T$ or some trailing average of them.
We will mention some specific examples in \S\ref{s-time-series},
where we describe some well known covariance predictors in
the time series context.

As a specific example, $y_i$ is the vector of returns of 
$n$ financial assets over day $i$,
with mean small enough to ignore, or already subtracted.
The return vector $y_i$ is not known on day $i$.
The feature vector $x_i$ includes quantities known on day $i$,
such as economic indicators, volatility indices, previous realized 
trading volumes or returns, and (the entries of) $y_{i-1}y_{i-1}^T$.
The predicted covariance $\hat \Sigma(x_i)$ can be interpreted
as a (time-varying) risk model that depends on the features.

\subsection{Parametrizing and fitting covariance predictors}
\label{s-para-fit}
Covariance predictors can range from simple,
such as the empirical covariance of the outcome on the training 
data (which is constant, \ie, does not depend on the features),
to very complex, for example a neural network or a decision tree
that maps the feature vector to a positive definite matrix.
(We will describe many covariance predictors 
in \S\ref{s-previous-work}.)
Many predictors include parameters that are chosen
by solving an optimization problem, typically
maximizing the log-likelihood on the training data, minus
some regularization.
In most cases this optimization problem is not convex,
so we have to resort to heuristic methods that approximately
solve it.   In a few cases, including our proposed method,
the fitting problem is convex, which means it can be reliably globally
solved.

In this paper we focus on a predictor that has the same form as a 
generalized linear model, \ie,
an affine function of the features followed by a function interpretable as 
an inverse link function
that maps vectors to symmetric positive definite matrices.
The associated fitting problem is 
convex, and so readily solved globally.

\subsection{Outline}
In \S\ref{s-whitening} we observe that covariance prediction is 
equivalent to finding a feature-dependent linear mapping 
of the outcome that whitens it, \ie, maps its distribution to
$\mathcal N(0,I)$.
Our proposed covariance predictor is based directly on this 
observation.
The interpretation also suggests that multiple covariance prediction
methods can be iterated, which we will see in examples
leads to improved performance.
In \S\ref{s-previous-work} we outline the large body of related previous
work.
We present our proposed method, the regression whitener, 
in \S\ref{s-inverse_cholesky_model}, and give
some variations and extensions of the method in \S\ref{s-variations}.
In \S\ref{s-factors} and \S\ref{s-ml-resid}
we illustrate the ideas and our method on two examples,
a financial time series and a machine learning residual problem.

\section{Feature-dependent whitening}\label{s-whitening}
In this section we show that the covariance prediction problem is
equivalent to finding a feature-dependent linear transform of the 
data that whitens it, \ie, results (approximately) in an $\mathcal N(0,I)$
distribution.

Given a covariance predictor $\hat \Sigma$,
define $L:\mathcal X \to \mathcal L$ as
\[
L(x) = \chol(\hat \Sigma(x)^{-1}) \in \mathcal L,
\]
where $\chol$ denotes the Cholesky factorization, and
$\mathcal L$ is the set of $n \times n$ lower triangular matrices with
positive diagonal entries.
For $A \in \symm_{++}^n$, 
$L=\chol(A)$ is the unique $L \in \mathcal L$ that satisfies $LL^T=A$.
Indeed, $\chol:\symm_{++}^n \to \mathcal L$ is a bijection, with inverse
mapping $L \mapsto LL^T$.
We can think of $L$ as a feature-dependent linear whitening 
transformation for the outcome $y$,
\ie, $z = L(x)^Ty$ should have an $\mathcal N(0,I)$ distribution.

Conversely we can associate with any feature-dependent 
whitener $L: \mathcal X \to \mathcal L$
the covariance predictor 
\[
\hat \Sigma (x) = (L(x)L(x)^T)^{-1} = L(x)^{-T}L(x)^{-1}.
\]
The feature-dependent whitener is just another parametrization 
of a covariance predictor.

\paragraph{Interpretation of Cholesky factors.}
Suppose $y \sim \mathcal N(0, (LL^T)^{-1})$,
with $L\in \mathcal L$.
The coefficients of $L$ are closely connected to the prediction
of $y_i$ (here meaning the $i$th component of $y$) from $y_{i+1}, \ldots, y_n$.
Suppose the coefficients $a_{i,i+1},\ldots,a_{i,n}$ minimize the mean-square error
\[
\Expect\left(y_i - \sum_{j=i+1}^n a_{i,j}  y_j\right)^2.
\]
Let $J_i$ denote the minimum value, \ie, the minimum mean-square error (MMSE)
in predicting $y_i$ from $y_{i+1}, \ldots, y_n$.
We can express the entries of $L$ in terms of the coefficients $a_{i,j}$ and $J_i$.

We have $L_{ii} = 1/\sqrt{J_{i}}$, \ie,
$L_{ii}$ is the inverse of the standard deviation of the prediction error.
The lower triangular entries of $L$ are given by
\[
L_{ji} = -L_{ii}a_{i,j}, \quad j=i+1,\ldots,n.
\]
This interpretation has been noted in several places, \eg, \cite{pourahmadi2011covariance}.
An interesting point was made in \cite{wu2003nonparametric}, namely that an approach like this ``reduces the difficult and nonintuitive task of modelling a covariance matrix
to the more familiar task of modelling $n-1$ regression problems''.

\paragraph{Log-likelihood.}
For future reference, we note that the 
log-likelihood of the sample $x,y$ with whitener
$L(x)$, and associated covariance predictor
$\hat \Sigma(x) = L(x)^{-T}L(x)^{-1}$, is
\BEA
\label{e-log-like-L}
\lefteqn{
-(n/2) \log (2\pi) - (1/2)\log \det \hat \Sigma(x)
- (1/2)y^T \hat \Sigma(x) ^{-1} y}
\nonumber\\
&=&
-(n/2)\log (2\pi) + \sum_{j=1}^n \log L(x)_{jj} 
-(1/2) \| L(x)^T y \|_2^2.
\EEA
The log-likelihood function \eqref{e-log-like-L} is 
a concave function of $L(x)$ \cite{boyd_convex_2004}.

\subsection{Iterated covariance predictor}\label{s-iterated-predictor}
The interpretation of a covariance predictor as a feature-dependent whitener 
leads directly to the idea of iterated whitening.
We first find a whitener $L_1$, to obtain the (approximately) 
whitened data $z_i^{(1)} = L_1(x_i)^T y_i$.
We then find a whitener $L_2$ for the data $z_i^{(1)}$ to obtain
$z^{(2)}_i = L_2(x_i)^T z^{(1)}_i$.  We continue this way $K$ iterations 
to obtain our final whitened data
\[
z^{(K)}_i = L_K(x_i)^T \cdots L_1(x_i)^T y_i.
\]
This composition of $K$ whiteners is the same as the whitener
\[
L(x) = L_1(x) \cdots L_K(x) \in \mathcal L,
\]
with associated covariance predictor
\[
\hat \Sigma (x) = L_1(x)^{-T} \cdots L_K(x)^{-T} L_K(x)^{-1} \cdots L_1(x)^{-1}.
\]

The log-likelihood of sample $x,y$ for this iterated predictor is
\[
-(n/2)\log (2\pi) + \sum_{k=1}^K \sum_{j=1}^n \log (L_k(x))_{jj} 
-(1/2) \| L_K(x)^T \cdots L_1(x)^T y \|_2^2.
\]
This function is multi-concave, \ie, concave in each $L_k(x)$,
with the others held constant,
but not jointly in all of them.

Our examples will show that iterated whitening can improve
the performance of covariance prediction over that of the 
individual whiteners.
Iterated whitening as related to the concept of boosting in
machine learning, where a number of weak learners are applied in succession
to create a strong learner \cite{freund1996experiments}.  

\section{Previous work}
\label{s-previous-work}

In this section we review the very large body of previous and related 
work, which goes back many years, using the notation of this paper when possible.

\subsection{Heteroscedasticity}
The ordinary linear regression method assumes the residuals have constant variance.
When this assumption is violated, the data or model is said to be heteroscedastic, meaning that the variance of the errors depends on the features.
There exist a number of tests to check whether this exists in a dataset \cite{anscombe1961examination,cook1983diagnostics}.
A common remedy for heteroscedasticity once it is identified is to apply an invertible function to the outcome (when it is positive), \eg, $\log(y)$, $1/y$, or $\sqrt{y}$, to make the variances more constant or homoscedastic \cite{davidian1987variance}.
In general, one needs to fit a separate model for the (co)variance of the prediction residuals, which can be done in a heuristic way by doing a linear regression on the absolute or squared value of the residuals \cite{davidian1987variance}.

\subsection{General examples}
Here we describe some of the covariance predictors that have been
proposed.  We focus on how the predictors are parametrized and fit
to training data, noting when the fitting problem is convex.
In some cases we report on methods originally developed
for fitting a constant covariance, but are readily adapted to fitting
a covariance predictor.

\paragraph{Constant predictor.}
The simplest covariance predictor is constant, \ie, $\hat \Sigma(x) = \Sigma$
for some $\Sigma \in \symm_{++}^n$.
The simplest way to choose $\Sigma$ given training data is the empirical 
covariance, which maximizes the log-likelihood of the training data.
More sophisticated estimators of a constant covariance employ 
various types of regularization \cite{friedman2008sparse}, 
or impose special structure on $\Sigma$, 
such as being diagonal plus low rank \cite{rubin1982algorithms}
or having a sparse inverse \cite{dempster1972covariance,friedman2008sparse}.
Many of these predictors are fit by solving a convex 
optimization problem,
with variable $\hat \Sigma(x)^{-1}$, the precision matrix.

\paragraph{Diagonal predictor.}
Another simple predictor has diagonal covariance, of the form 
\BEQ
\hat \Sigma(x) = \diag(\exp(Ax+b)),
\label{eq:diagon_model}
\EEQ
where $A$ and $b$ are the predictor parameters, and $\exp$ is elementwise.
With this predictor the log-likelihood is a 
concave function of $A$ and $b$.
(In fact, the log-likelihood is separable across the rows of $A$ and $b$.)
Fitting a diagonal predictor
by maximizing log-likelihood (minus a convex regularizer) is
a convex optimization problem.
This is a special case of Pourahmadi's $LDL^T$ approach 
where $L=I$ \cite{pourahmadi1999joint}.
This covariance predictor for (the special case of) $n=1$ was
implemented in the R package \verb|lmvar|
\cite{lmvar}.

This simple diagonal predictor can be used in an iterated covariance 
predictor, preceded by a constant whitener.
For example, we
start with a constant base covariance $\Sigma^\text{const}$,
with eigenvalue decomposition $\Sigma^\text{const} = U \diag(\lambda) U^T$.
The iterated predictor has the form
\[
\hat \Sigma(x) = U \diag( \lambda \circ \exp(Ax+b)) U^T,
\]
where $\circ$ denotes the elemementwise (Hadamard) product,
and $A$ and $b$ are our predictor parameters.  
In this covariance prediction model we fix the eigenvectors of 
the (base, constant) covariance, and scale the eigenvalues based on the features.
Fitting such a predictor is a convex optimization problem.

\paragraph{Cholesky and $LDL^T$ predictors.}
Several authors use the Cholesky parametrization of positive 
definite matrices or the
closely related $LDL^T$ factorization of $\Sigma$ or $\Sigma^{-1}$.
Perhaps the first to do this was Williams, who in 1996 proposed making the output of a neural network the lower triangular entries and logarithm of the diagonals of the Cholesky decomposition of the inverse covariance matrix \cite{williams1996using}.
He used the log-likelihood as the objective to be maximized, and provided the partial derivatives of the objective with respect to the network outputs.
The associated fitting problem is not convex.
Williams's original work was repeated without citation in \cite{dorta2018structured}.
William's original work was expanded upon and interpreted by Pourahmadi in a series of 
papers \cite{pourahmadi1999joint,pourahmadi2000maximum}; a good summary of 
these papers can be found in \cite{pourahmadi2011covariance}.
A number of regularization functions for this problem have been considered in \cite{huang2006covariance}.
However, to the best of the authors' knowledge, none of these problems is convex, 
but some are bi-convex, for example Pourahmadi's $LDL^T$ formulation, 
which has a log-likelihood that is concave in $L$,
and also in the variables $\log D_{ii}$,
but not in both sets of variables.
Some of the aforementioned methods have been implemented in the R package \verb|jmcm| \cite{jmcm}.

\paragraph{Linear covariance predictors.}
Some methods proposed by researchers to fit a constant covariance can be 
readily extended to fit a covariance predictor, \ie, one 
that depends on the feature vector $x$.
For example, the linear covariance model \cite{anderson1973asymptotically} 
fits a constant covariance
\BEQ\label{e-lin-cov-pred}
\hat \Sigma = \sum_{k=1}^K \alpha_k \Sigma_k,
\EEQ
where $\Sigma_1,\ldots,\Sigma_K$ are known symmetric matrices
and $\alpha_1,\ldots,\alpha_K\in\reals$ are coefficients that are fit to data.
Of course, $\alpha_1,\ldots,\alpha_K$ must be chosen such that
$\Sigma$ is positive definite; a sufficient condition, 
when $\Sigma_k\in \symm_{++}^n$, is $\alpha \geq 0$, $\alpha \neq 0$.
This form is readily extended to be a covariance predictor by making 
the coefficients $\alpha_i$ functions of $x$, for example 
affine, $\alpha = Ax+b$, where $A$ and $b$ are model parameters.
(We ignore here the constraint on $\alpha$, discussed in \S\ref{s-pos-diag}.)
For this linear parametrization of the covariance matrix, the 
log-likelihood is not a concave function of $\alpha$, so fitting
such a predictor is not a convex optimization problem.

Using the inverse covariance or precision matrix, the natural
parameter in the exponential family representation of a Gaussian distribution,
we do obtain a concave log-likelihood function.
The model for a constant covariance is
\[
\hat \Sigma = \left( \sum_{k=1}^K \alpha_k \theta_k \right)^{-1}
\]
where $\theta_1,\ldots,\theta_K\in\symm_{++}^n$, and $\alpha_k$ are the 
parameters to be fit, with the constraint $\alpha \geq 0$, $\alpha \neq 0$.
The log-likelihood for a sample $x, y$ is
\[
-n \log (2\pi) + \log \det\left(\sum_{k=1}^K \alpha_i \theta_i\right) - y^T 
\left(\sum_{k=1}^K \alpha_k \theta_k\right) y,
\]
which is a concave function of $\alpha$.
This model is readily extended to give a covariance predictor
using $\alpha = Ax+b$, where $A$ and $b$ are model parameters.
(Here too we must address the issue of the constraints on $\alpha$.)
Fitting the parameters $A$ and $b$ is a convex optimization problem.

Several methods can be used to find a suitable basis $\Sigma_1, \ldots, \Sigma_K$
or $\theta_1, \ldots, \theta_K$.
For example, we can run a $k$-means like algorithm that alternates between
assigning data points to the covariance or preicision matrix that has highest
likelihood, and updating each matrix by maximizing likelihood (possibly minus 
a regularizer) using the data assigned to it.

\paragraph{Log-linear covariance predictors.}
Another example of a constant covariance method that can be readily extended 
to a covariance predictor is the log-linear covariance model.
The 1992 paper by Leonard and Hsu \cite{chiu1996matrix}
propose using the matrix exponential, which maps the vector space $\symm^n$ 
($n \times n$ symmetric matrices) onto $\symm_{++}^n$,
for the purpose of fitting a constant covariance matrix.
To extend this to covariance prediction, we take
\[
\hat \Sigma(x) = \exp Z(x), \qquad Z(x) = Z_0 + \sum_{i=1}^m x_i Z_i,
\]
where $Z_0, \ldots, Z_m$ are (symmetric matrix) model parameters.

The log-likelihood for a sample $x, y$ is
\[
-n \log (2\pi) - \Tr Z(x) - y^T \left( \exp Z(x)\right)^{-1} y,
\]
which unfortunately is not concave in the parameters.
In a 1999 paper, Williams proposed using the matrix exponential as the 
final layer in a neural network that 
predicts covariances \cite{williams1999matrix}, \ie, the neural
network maps $x$ to (the symmetric matrix) $Z(x)$.

\paragraph{Hard regimes and modes.}
Covariance predictors can be built from a finite number of given
covariance matrices, $\Sigma_k$, $k=1,\ldots, K$.
The index $k$ is often referred to as a (latent, unobserved) 
mode or regime.
The predictor has the form
\[
\hat \Sigma(x) = \Sigma_{k}, \qquad k = \phi(x),
\]
where $\phi: \mathcal X \to \{1,\ldots, K\}$ is a
$K$-way classifier (tuned with some parameters).
We do not know a parametrization of classifiers for which
the log-likelihood is concave, but there are several heuristics
that can be used to fit such a model.
This regime model is a special case of a linear covariance predictor
described above,
when the coefficients $\alpha$ are restricted to be unit vectors.

One method proceeds as follows.
Given the matrices $\Sigma_1, \ldots, \Sigma_K$,
we assign to each data sample $x,y$ 
the value of $k$ that maximizes the likelihood, \ie, the regime
that best explains it.
We then fit the classifier $\phi$ to the data pairs $x,k$.
When the classifier is a tree,
we obtain a covariance tree, with each leaf associated with one
of the regime covariances \cite{covariance_trees}.

To also fit the regime covariance matrices $\Sigma_1, \ldots, \Sigma_K$, we fix
the classifier, and then fit each $\Sigma_k$ to the data points
with $\phi(x)=k$.
This procedure can be iterated, analogous to the $k$-means algorithm.

\paragraph{Soft regimes or modes.}
We replace the hard classifier described above with a
soft classifier 
\[
\phi : \mathcal X \to \{\pi \in \reals^K \mid \pi \geq 0, ~\ones^T \pi =1\}.
\]
(We can interpret $\pi_k$ as the probability of regime $k$,
given $x$.)
We form our prediction as a mixture of the (given) regime precision matrices,
\[
\hat \Sigma(x) = \left(
\sum_{k=1}^K \phi(x)_k \Sigma_k^{-1} \right)^{-1},
\]
(which has the same form as a linear covariance predictor with precision
matrices, described above).
With this predictor, the log-likelihood is a concave
function of $\phi(x)$, so when $\phi$ is an affine function of $x$,
\ie, $\phi(x)= Ax+b$, the fitting problem is convex.
(Here we have $\ones^T b =1$ and $\ones^TA=0$, which implies 
that $\ones^T\phi(x) = 1$ for all $x$, and we ignore the
issue that we must have $Ax+b \geq 0$, which we address in
\S\ref{s-pos-diag}.)

A more natural soft predictor is multinomial logistic regression, with
\[
\phi(x) = \frac{\exp q}{\ones^T \exp q}, \qquad
q = Ax+b,
\]
where $A$ and $b$ are parameters.
With this parametrization, the log-likelihood is not concave in 
$A$ and $b$, so fitting such a predictor is not 
a convex optimization problem.

\paragraph{Laplacian regularized stratified covariance predictor.}
Laplacian regularized stratified models, described in
\cite{tuck2019distributed,tuck2021portfolio,tuck2020fitting}, can be used
to develop a covariance predictor.
To do this, one bins $x$ into $K$ categories, and gives a
(possibly different) covariance matrix for 
each of the $K$ bins.  (This is the same as a hard regime model with
the binning serving as 
a very simple classifier that maps $x$ to $\{1, \ldots, K\}$.)
The predictor is parametrized by the 
covariance matrices $\Sigma_1, \ldots, \Sigma_K$; the log-likelihood
is concave in the precision matrices
$\Sigma_1^{-1}, \ldots, \Sigma_K^{-1}$.
From the log-likelihood we subtract a
Laplacian regularizer that encourages the precision matrices 
associated with neighboring bins to be close.
Fitting such a predictor is a convex optimization problem.

\paragraph{Local covariance predictors.}
We mention one more natural covariance predictor, based on the idea
of a local model \cite{cleveland1988}.  We describe here a simple version.
The predictor uses the full set of training data, $x_i,y_i$, $i=1, \ldots, N$.
The covariance predictor is
\[
\hat \Sigma(x) = \sum_{i=1}^N \alpha_i y_iy_i^T, \qquad \alpha_i =
\frac{\phi(\|x-x_i\|_2)}{\sum_{j=1}^N \phi(\|x-x_j\|_2)},
\]
where $\phi: \reals_+ \to \reals_{++}$ 
is a radial kernel function.  The most common choice is the Gaussian kernel,
$\phi(u)=\exp(-u^2/\sigma^2)$,
where $\sigma$ is a (characteristic distance) parameter.
(One variation is to take $\alpha_i = 1/K$ for the $K$-nearest neighbors of $x$
among $x_1, \ldots, x_N$, and zero otherwise, with $K \>n$.)
We recognize this as a special case of a 
linear covariance predictor \eqref{e-lin-cov-pred},
with a specific choice of the mapping from $x$ to the coefficients,
and $\Sigma_i = y_yy_i^T$.

\subsection{Time series covariance forecasters}\label{s-time-series}
Here we assume that $i$ denotes time period or epoch.
At time $i$, we know the previous realized values 
$y_{i-1}, y_{i-2}, \ldots$, so functions of them can appear 
in the feature vector $x_i$.  We write $\hat \Sigma(x_i)$
as $\hat \Sigma_i$.

\paragraph{SMA.}
Perhaps the simplest covariance predictor for a time series
(apart from the constant predictor) is the simple moving average (SMA)
predictor, which averages $M$ previous values of $y_iy_i^T$ to 
form $\hat \Sigma_i$,
\[
\hat \Sigma_i = \frac{1}{M}\sum_{j=1}^{M} y_{i-j} y_{i-j}^T.
\]
Here $M$ is called the memory of the predictor.
The SMA predictor follows the recursion
\[
\hat \Sigma_{i+1} = \hat \Sigma_i + 
\frac{1}{M}(y_i y_i^T - y_{i-M}y_{i-M}^T).
\]
Fitting an SMA predictor does not explicitly involve solving
a convex optimization problem, but it does maximize the 
(concave) log-likelihood of the observations $y_{i-j}$, $j=1, \ldots, M$.

\paragraph{EWMA.}
The exponentially 
weighted moving average (EWMA) predictor uses exponentially weighted
previous values of $y_i y_i^T$ to form $\hat \Sigma_i$,
\BEQ\label{e-ewma}
\hat \Sigma_i = \alpha_i \sum_{j=1}^{i-1} \gamma^{i-j}
 y_j y_j ^T, \qquad \alpha_i = \left( \sum_{j=1}^{i-1}
\gamma^j \right)^{-1},
\EEQ
where $\gamma \in (0,1]$ is the forgetting factor, 
often specified by the half-life $T^\text{half} = - (\log 2)/(\log \gamma)$
\cite{hawkins2008multivariate,harper2009exploring}.
This predictor follows the recursion
\[
\hat \Sigma_{i+1} = \gamma \frac{\alpha_{i+1}}{\alpha_i}\hat \Sigma_i + 
\alpha_{i+1} y_i y_i^T.
\]
The EWMA covariance predictor is widely used in finance
\cite{longerstaey1996riskmetricstm,menchero2011barra}.
Like SMA, the EWMA predictor maximizes a (concave) weighted likelihood of
past observations.

\paragraph{ARCH.}
The autoregressive conditional heteroscedastic (ARCH) predictor
\cite{engle1982autoregressive} is a variance predictor (\ie, $n=1$)
that uses features $x_i=(y_{i-1}^2,\ldots,y_{i-M}^2)$ and has the form
\[
\hat \Sigma_i = \alpha_0 + \sum_{j=1}^M \alpha_j y_{i-j}^2,
\]
where $\alpha_j \geq 0$, $j=0,\ldots,M$,
and $M$ is the memory or order of the predictor.
In the original paper on ARCH, Engle also suggested that 
external regressors could be used as well to predict the variance,
which is readily included in the predictor above.
A one-dimensional SMA model is a special case of an 
ARCH model with $\alpha_0=0$ and $\alpha_j=1/M$.
The log-likelihood is not a concave function of the parameters $\alpha_0,
\ldots, \alpha_M$, so fitting an ARCH predictor requires 
solving a nonconvex optimization problem.

\paragraph{GARCH.}
The generalized ARCH (GARCH) \cite{bollerslev1986generalized} model, 
originally introduced by Bollerslev, is a generalization of ARCH 
that includes prior predicted values of the variance in the features.
It has the form
\[
\hat \Sigma_i = \alpha_0 + \sum_{j=1}^M \alpha_j y_{i-j}^2 
 + \sum_{j=1}^M \beta_j \hat \Sigma_{i-j},
\]
where $\alpha_i$ and $\beta_i$ are nonnegative parameters.
The SMA and EWMA models with $n=1$ are both special cases of 
a GARCH model.
Like ARCH, the log-likelihood function for the GARCH model is not 
concave, so fitting it requires solving a nonconvex optimization problem.

\paragraph{Multivariate GARCH.}
While the original GARCH model is for one-dimensional $y_i$,
it has been extended to multivariate time series.
For example, the diagonal GARCH model \cite{bollerslev1988capital} uses a separate GARCH model for each entry of $y_i$,
the constant correlation GARCH model \cite{bollerslev1990modelling} assumes a constant correlation between the entries of $y_i$,
and the BEKK model (named after Babba, Engle, Kraft, and Kroner) \cite{engle1995multivariate} is a generalization of all of the above models.
There exist many other GARCH variants (see, \eg, \cite{francq2019garch} and the references therein).
None of these predictors have a concave log-likelihood,
so fitting them involves solving a nonconvex optimization problem.

\paragraph{Time-varying factor models.}
In a covariance factor model, 
we regress $y_i$ on some factors $w_i$, perhaps with exponential weighting,
to get $y_i = F_i w_i + \epsilon_i$
\cite[\S3]{grinold2000active}.
Assuming $w_i \sim \mathcal N(0,
\Sigma^\text{fact})$ and $\epsilon_i \sim \mathcal N(0,\diag(d_i))$,
leads us to the time-varying factor covariance model \cite{su2017time}
\[
\hat \Sigma_i = F_i\Sigma^\text{fact} F_i^T + \diag(d_i).
\]
The methods of this paper can be used form a covariance predictor 
for the factors, which can also depend on features.

\paragraph{Hidden Markov regime models.}  Hard regime models can be used 
in the context of time series, with a Markov model for the 
transitions among regimes \cite{bilmes1998gentle}.
One form for this predictor estimates the probability distribution 
of the current latent state or regime, and then forms 
a weighted sum of the precision matrices as our estimate.

\section{Regression whitener}
\label{s-inverse_cholesky_model}

In this section we describe a simple feature-dependent whitener,
in which $L(x)$ is an affine function of $x$,
\[
\diag(L(x)) = Ax + b, 
\qquad \mathbf{offdiag}(L(x)) = Cx + d,
\]
where $\diag$ gives the vector of diagonal entries,
and $\mathbf{offdiag}$ gives the strictly lower triangular 
entries in some fixed order.
The regression model coefficients are
\[
A\in\reals^{n \times p}, \quad b\in\reals^n,
\quad  C\in\reals^{k\times p}, \quad d\in\reals^{k},
\]
with $k=n^2/2-n/2$ denoting the number of strictly lower triangular 
entries of an $n \times n$ matrix.
The total number of parameters in our model is
\BEQ\label{e-total-params}
np+n+kp+k = \frac{n(n+1)}{2}(p+1).
\EEQ
Our model parameters can be assembled into a single 
$\frac{n(n+1)}{2} \times (p+1)$ parameter matrix
\[
P = \left[ \begin{array}{cc} A & b \\ C & d \end{array}\right].
\]
The top $n$ rows of $P$ give the diagonal of $L$; its last 
column gives the constant or offset part of the model, \ie, $L(0)$.

The log-likelihood of the regression whitener is a concave 
function of the parameters $(A,b,C,d)$.  
We have already noted that the log-likelihood 
\eqref{e-log-like-L} is a concave function of $L(x)$, which in
turn is a linear function of the parameters $(A,b,C,d)$.
The composition of a concave function and a linear function is 
concave and the sum (over the training samples) preserves concavity
\cite[\S 3.2]{boyd_convex_2004}.

With the regression whitener,
the precision matrix $\hat \Sigma(x)^{-1} = L(x)L(x)^T$ is
a quadratic function of the feature vector $x$; its inverse,
the covariance $\hat \Sigma(x)$, is a more complex function 
of $x$.

\subsection{The issue of positive diagonal entries}\label{s-pos-diag}
To have $L(x) \in \mathcal L$ for all $x\in \mathcal X$,
we need $Ax+b >0$ (elementwise) for all $x\in \mathcal X$.
When $\mathcal X= \reals^p$, this holds only when $A=0$
and $b>0$.  Such a whitener, which has fixed diagonal entries but 
lower triangular entries that can depend on $x$, can still have value,
but this is a strong restriction.
The condition that $Ax+b>0$ for all $x\in \mathcal X$ is convex 
in $(A,b)$, and leads to a tractable constraint on these coefficients
for many choices of the feature set $\mathcal X$.

\paragraph{Box features.}
Perhaps the simplest case is $\mathcal X=
\{ x \mid \|x\|_\infty \leq 1\}$, \ie, the unit box.
This means that \emph{all features lie between $-1$ and $1$}.
This can be ensured in several reasonable ways.  First, we can simply clip or 
Winsorize our raw features $\tilde x$, using 
\[
x= \clip(\tilde x) = \min\{1, \max\{-1,\tilde x\}\}
\]
(interpreted elementwise).
Another reasonable approach is to map the values of $\tilde x_i$ (the $i$th 
component of $x$) into $[-1,1]$,
for example, by taking $x_i = (2) \quantile_i(\tilde x_i)-1$, 
where $\quantile_i(\tilde x_i)$ is the quantile of $\tilde x_i$.
Another approach is to scale the values of $\tilde x_i$ by its 
minimum and maximum by taking
\[
x_i = 2\frac{\tilde x_i - m_i}{M_i-m_i} - 1,
\]
where $m_i$ and $M_i$ are the
smallest and largest values (elementwise) of $x_i$ in the training data.
(When this is used on data not in the training set we would also clip 
the result of the scaling above to $[-1,1]$.)
We will assume from now on that $\mathcal X= \{x \mid 
\|x\|_\infty \leq 1\}$.

As a practical matter we work the non-strict inequality
$Ax+b \geq \epsilon$ for all $x\in \mathcal X$, where
$\epsilon>0$ is given, and the inequality is meant elementwise.
The requirement that $Ax+b\geq \epsilon$ for all $\|x\|_\infty \leq 1$
is equivalent to
\BEQ\label{e-diag-constr}
\|A\|_{\text{row},1} \leq b - \epsilon,
\EEQ
where $\|A\|_{\text{row},1} \in \reals_+^n$
is the vector of $\ell_1$ norms of the rows of $A$, \ie,
\[
\left(\|A\|_{\text{row},1}\right)_i = \sum_{j=1}^p |A_{ij}|.
\]
The constraint \eqref{e-diag-constr} is a convex (polyhedral) 
constraint on $(A,b)$.

\subsection{Fitting}
Consider a training data set $x_1,\ldots,x_N$, $y_1,\ldots,y_N$.
We will choose $(A,b,C,d)$ to maximize the log-likelihood of 
the training data, 
minus a convex regularizer $R(A,b,C,d)$,
subject to the constraint \eqref{e-diag-constr}.

This leads to the convex optimization problem
\BEQ
\begin{array}{ll}
\text{maximize} & (1/N)\sum_{i=1}^{N}\left( \sum_{j=1}^n \log (L_i)_{jj} - (1/2)
\|L_i^Ty_i\|_2^2 \right) - R(A,b,C,d)\\
\text{subject to} & \mathbf{diag}(L_i) = Ax_i + b, \quad i=1,\ldots,N,\\
& \mathbf{offdiag}(L_i) = Cx_i + d, \quad i=1,\ldots,N,\\
& \|A\|_{\text{row},1}\leq b - \epsilon,
\end{array}
\label{eq:prob1}
\EEQ
with variables $A$, $b$, $C$, $d$.
We note that the first term in the objective guarantees 
that $L(x)_{jj} > 0$ for 
all training feature values $x=x_i$; 
the last (and stronger) constraint ensures that
$L(x)_{jj} \geq \epsilon$ for \emph{any} feature vector in $\mathcal X$,
\ie, $\|x\|_\infty \leq 1$.

\subsection{Regularizers}
There are many useful regularizers for the covariance prediction
problem \eqref{eq:prob1}, a few of which we mention here.

\paragraph{Trace inverse regularization.}
Several standard regularizers used in covariance fitting can be included in $R$.
For example trace regularization of the precision matrix, on the 
training data, is given by
\[
\lambda \frac{1}{N} \sum_{i=1}^N \Tr \hat \Sigma(x_i)^{-1},
\]
where $\lambda >0$ is a hyper-parameter.
This can be expressed in terms of our coefficients as
\[
R(A,b,C,d) =
\lambda \frac{1}{T} \sum_{t=1}^T \|L_i\|_F^2,
\]
\ie, $\ell_2$-squared regularization on $L_i$.
This can be expressed directly in terms of $A,b,C,D$ as
\[
\lambda \frac{1}{T} \sum_{t=1}^T \left( \|Ax_i + b\|_2^2+
\|Cx_i + d\|_2^2 \right).
\]

We can simplify this regularizer, and remove its dependence on the training
data, by assuming that the entries of the features are approximately independent and 
uniformly distributed on $[-1,1]$.  This leads to the approximation (dropping a 
constant term)
\[
R(A,b,C,d) = 
\lambda \frac{n}{12} \left\| \left[ \begin{array}{c} A \\ C \end{array} \right] \right\|_F^2.
\]
This exactly the traditional ridge or quadratic regularizer on 
the model coefficients, not including the offset.

\paragraph{Feature selection.}
The regularizer
\[
R(A,b,C,d) = \lambda \sum_{i=1}^p \| (a_i,c_i) \|_2,
\]
where $a_i$ and $c_i$ are the $i$th columns of $A$ and $C$,
is the sum of the norms of the first $p$ columns of $P$.
This is a well-known sparsifying regularizer, that tends to give 
coefficients with $(a_i,c_i)=0$, for many values of $i$ \cite{meier2008group}.
This means that the feature entry $x_i$ is not used in the model.

\paragraph{Dual norm regularization.}
The total number of parameters in our model, given by \eqref{e-total-params},
can be quite large if $n$ is moderate or $p$ is large.
An interesting regularizer that leads to a more interpretable covariance
predictor can be obtained with dual norm regularization (also called
trace or nuclear norm regularization),
\[
\lambda \left\|\begin{bmatrix}A\\C\end{bmatrix}\right\|_*,
\]
where $\|\cdot \|_*$ is the dual of the $\ell_2$ norm of a matrix,
\ie, the sum of the singular values,
and $\lambda>0$ is a hyper-parameter.
This regularizer is well known to encourage its argument to be low rank
\cite{vandenberghe1996semidefinite, recht2010guaranteed}.

When $\begin{bmatrix}A \\C\end{bmatrix}$ is (say) rank $r$,
it can be expressed as the product of two smaller matrices,
\[
\begin{bmatrix}
A \\ C
\end{bmatrix} = UV,
\]
where $U\in\reals^{n + k\times r}$ and $V\in\reals^{r \times p}$.
For notational convenience, we let $L^i = \mathbf{mat}(U_i)$, where $U_i$ is the $i$th column of $U$
and $\mathbf{mat}:\reals^{n+k}\to\mathcal L$ takes the diagonal and strictly
lower triangular entries and gives the corresponding lower triangular
matrix in $\mathcal L$.
We also let $L^0 = \mathbf{mat}(b,d)$.
Finally, we let $V_i$ denote the $i$th row of $V$.

With this low rank coefficient matrix, the process of prediction can 
be broken down into two simple steps.
We first compute $r$ latent factors $l_i = V_i^Tx_i$,
$i=1,\ldots,r$, which are linear in the features.
Then $L(x)$ is a sum of $L^0,\ldots,L^r$, weighted by $l_1,\ldots,l_r$,
\[
L(x) = L^0 + \sum_{i=1}^r l_i L^i.
\]
Thus our whitener $L(x)$ is always a linear combination of 
$L^0,\ldots,L^r$.

\subsection{Ordering and permutation} \label{s-ordering}
The ordering of the entries in the data $y$ matters.
That is, if we fit a model $\hat \Sigma_1$ with training data $y_i$,
then fit another model $\hat \Sigma_2$ to $Qy_i$,
where $Q$ is a permutation matrix, we generally do not have
$\hat \Sigma_1(x) = Q^T\hat \Sigma_2(x)Q$.
This was noted previously in \cite[\S2.2.4]{pourahmadi2011covariance},
where the author states that ``the factors of the Cholesky decomposition are
dependent on the \emph{order} in which the variables appear in the random 
vector $y_i$''.
This has been noted as a pitfall of Cholesky-based approaches and 
can lead to significant differences in forecast performance \cite{heiden2015pitfalls},
although on problems with real data we have not observed large differences.

This dependence of the prediction model on the ordering of the 
entries of $y$ is unattractive, at least theoretically.
It also immediately raises the question of how to choose a good ordering for
the entries of $y$.
We have observed only small differences in the performance of 
covariance predictors obtained by permuting the entries of $y$, so 
perhaps this is not an issue in practice.
A reasonable approach is to order the entries in such a way that 
correlated entries (say, under a base constant model) are near
each other \cite{rothman2010new}.
But we consider the question of how to order the variables in a 
regression whitener to be an open question.

There are a number of simple practical ways to deal with this issue.
One is to fit a number of models with different orderings of $y$,
and choose the model with the best out of sample likelihood,
just as we might do with regularization.
In this case we are treating the ordering as a hyper-parameter.

A practical method to obtain a model that is 
at least approximately invariant under ordering of the entries of $y$
is to fit a number of models $\hat \Sigma_1, \ldots, \hat \Sigma_K$ that
using different orderings, and then to fuse the models via
\[
\hat \Sigma(x) = \left(\frac{1}{K}\sum_{i=1}^K \hat \Sigma_i(x)^{-1}\right)^{-1}.
\]

Finally, we note that a permutation can be thought of as a very simple 
whitener in an iterated whitener.  It evidently does not whiten the data,
but when iterated whitening is done, the permutation can affect the performance 
of downstream whiteners, such as our regression
whitener, that depends on the ordering of the entries of $y$.

\subsection{Implementation}
Many methods can be used to solve the convex optimization 
problem \eqref{eq:prob1}.   Here we describe some good choices,
which are used in our implementation.

\paragraph{L-BFGS.}
We have observed that with reasonably chosen regularization,
the constraint $\|A\|_{\text{row},1} \leq b- \epsilon$ is rarely 
active at the solution of \eqref{eq:prob1}.
This suggests that we ignore the constraint, solve the problem,
and check at the end if it is active.
When the regularizer $R$ is differentiable,
the limited-memory Broyden Fletcher Golbfarb Shanno (L-BFGS) method
is well suited to solving this problem, after eliminating $L_i$.
The gradients of the objective with respect to $(A,b,C,d)$ are
straightforward to work out.

\paragraph{L-BFGS-B formulation.}
We can use L-BFGS-B (L-BFGS with box constraints)
\cite{liu_limited_1989} to efficiently solve the 
constrained problem \eqref{eq:prob1}, when $R$ is differentiable.
We reformulate it as the smooth box-constrained problem
\[
\begin{array}{ll}
\text{maximize} & (1/N) \sum_{i=1}^{N} \left( \ones^T\log 
(\mathbf{diag}(L_i)) - (1/2)\|L_i^Ty_i\|_2^2\right) - R(A,b,C,d)\\
\text{subject to} & \mathbf{diag}(L_i) = (A_+ - A_-)x_i + (A_+ + A_-)\ones + \epsilon + b_+, \quad i=1,\ldots,N,\\
& \mathbf{offdiag}(L_i) = Cx_i + d, \quad i=1,\ldots,N,\\
& A_+ \geq 0, \quad A_- \geq 0, \quad b_+ \geq 0,\\
& A= A_+-A_, \quad 
b = (A_+ + A_-)\ones + \epsilon + b_+,
\end{array}
\]
with variables $A_+$, $A_-$, $b_+$, $C$, $d$.
Here we have split $A$ into its positive and negative parts, and
take $b = (A_+ + A_-)\ones + \epsilon + b_+$.

\paragraph{Implementation.}
We have developed a Python-based object-oriented implementation of 
the ideas described in this paper, which is freely available online at
\[
\verb|www.github.com/cvxgrp/covpred|.
\]
The only dependencies are \verb|numpy| and \verb|scipy|,
and we use \verb|scipy|'s built-in LBFGS-B implementation.
The central object in the package is the \verb|Whitener| class,
which has three methods: \verb|fit|, \verb|whiten|, and \verb|score|.
The \verb|fit| method takes a training dataset given as \verb|numpy| 
matrices, and fits the parameters of the whitener.
The \verb|whiten| method takes a dataset
and returns a whitened version of the outcome as well as 
$L(x_i)$ and $\hat \Sigma(x_i)$ for each element of the dataset.
The \verb|score| method computes the log-likelihood of a dataset using the whitener.
The current implementation includes the following \verb|Whitener|s:
\begin{itemize}
\item \verb|ConstantWhitener|, a constant $\Sigma$.
\item \verb|DiagonalWhitener|, as described in \eqref{eq:diagon_model}.
\item \verb|SMAWhitener| and \verb|EWMAWhitener|,
as described in \S\ref{s-time-series}.
\item \verb|RegressionWhitener|, described in \S\ref{s-inverse_cholesky_model}.
\item \verb|PermutationWhitener|, permutes the entries in $y$ given a permutation.
\item \verb|IteratedWhitener|, described in \S\ref{s-whitening},
takes a list of whiteners, and applies them one by one.
\end{itemize}
These take arguments as appropriate, \eg, the memory for the SMA whitener,
and the choice of regularization for the regression whitener.
The examples we present later were implemented using this package, with the code
available in the \verb|examples| folder of the package linked above.

\section{Variations and extensions}\label{s-variations}
Here we list some variations on and extensions of the methods
described above.

\subsection{Multiple outcomes} \label{s-multiple-outcomes}
Each data record has the feature vector $x$ and a \emph{set} of outcomes,
possibly of varying cardinality.
(This reduces to our formulation when there is always just one outcome per
record.)
This is readily handled by simply replicating the data for each of 
the outcomes.  We transform the 
single record $x,y_1, \ldots, y_q$ into $q$ records of the form
$(x,y_1), \ldots, (x,y_q)$.  The methods described above can then be applied.
If $q$ can be large compared to $n$, it might be more efficient to transform
the data to outer products, \ie, replace the multiple outcomes
$y_1, \ldots, y_q$ into $Y = \sum_{i=1}^q y_iy_i^T$.

\subsection{Handling a nonzero mean} \label{s-nonzero-mean}
One simple extension is when the outcome vector $y$ has a nonzero mean,
and we model its distribution, conditioned on $x$, as $y \mid x \sim
\mathcal N (\hat \mu(x), \hat \Sigma(x))$.
One simple approach is sequential: 
we first fit a model $\hat \mu(x)$ of $y \mid x$,
for example by regression, subtract it from $y$ to create the 
prediction residuals or regression errors,
and then fit a covariance prediction to the residuals.

\paragraph{Joint prediction of conditional mean and covariance.}
It is also possible to handle the mean and covariance jointly, using
convex optimization.
With a nonzero mean $\hat \mu(x)$,
the log-likelihood \eqref{e-log-like-L} becomes
\[
-(n/2)\log (2\pi) + \sum_{j=1}^n \log L(x)_{jj} 
-(1/2) \| L(x)^T (y-\hat \mu(x))\|_2^2,
\]
where, as above, $L(x) = \chol(\hat \Sigma(x)^{-1})$.
This is concave in $L(x)$ and in $\hat \mu(x)$, but not jointly.

A change of variables, however, results in a 
jointly concave log-likelihood.
Changing the mean estimate variable $\hat \mu(x)$ to
$\nu(x) = L(x)^T \hat \mu(x)$, we obtain the log-likelihood function
\[
-(n/2)\log (2\pi) + \sum_{j=1}^n \log L(x)_{jj} 
-(1/2) \| L(x)^T y- \hat \nu(x)\|_2^2,
\]
which is jointly concave in $L(x)$ and $\nu(x)$.
We reconstruct the prediction of the mean and covariance of $y$ 
given $x$ as
\[
\hat \mu(x)=L(x)^{-T}\nu(x), \qquad 
\hat \Sigma(x)=L(x)^{-T}L(x)^{-1}.
\]
This trick is similar to, but not the same as, 
parametrizing a Gaussian
using the natural parameters in the exponential form,
$(\Sigma^{-1},\Sigma^{-1}\mu)$,
which results in a jointly concave log-likelihood function.
Our parametrization replaces the precision matrix $\Sigma^{-1}$ 
with its Cholesky factor $L$, and uses the parameters 
\[
(L,\nu) = (\chol(\Sigma^{-1}), \chol(\Sigma^{-1})^T\mu),
\]
but we still obtain a concave log-likelihood function.

To carry out joint mean and covariance prediction with the
regression whitener, we introduce two additional predictor
parameters $E\in \reals^{n\times p}$ and $f \in\reals^{n}$,
with $\nu(x) = Ex+f$.
Maximizing the log-likelihood minus a convex regularizer on $(A,b,C,d,E,f)$
is a convex problem, solved using the same methods as when
the mean of $y$ is presumed to be zero.

We note that while our prediction $\nu(x)$ is an affine 
function of $x$, our prediction of 
the mean $\hat \mu(x)$ is a nonlinear function of $x$.

\subsection{Structured covariance}
A few constraints on the inverse covariance matrix can be
expressed as convex constraints on $L$, and therefore
directly handled; others can be handled heuristically.
As an example, consider the constraint that $\hat \Sigma^{-1}$ be
banded, say, tri-diagonal.  This is equivalent to $L(x)$ having the same bandwidth
(and also, of course, being lower triangular), which in turn 
translates to rows of $C$ and $d$ corresponding to entries in $L$ outside the 
band being zero. This can be exactly handled by convex optimization.

Sparsity of $\hat \Sigma^{-1}$ (which corresponds to many pairs of 
the components of $y$ being independent, conditioned on all others) 
can be approximately handled by insisting that $L(x)$ be very sparse,
which in turn can be heuristically handled by using a regularization
that encourages row sparsity in $C$ and $d$, for example a 
sum of row norms.
Similar regularization functions have been used in the context of regularizing covariance predictors \cite{huang2006covariance}. 

\section{Example: Financial factor returns}\label{s-factors}
In this section we illustrate the methods described above on a
financial vector time series, where the outcome consists of
four daily returns, and the feature vector is constructed from
a volatility index as well as past realized volatilities.

\subsection{Outcome and features}

\paragraph{Outcome.}
We take $n=4$, with $y_i$ the daily returns of four Fama-French 
factors \cite{fama1992cross}:
\begin{itemize}
\item \emph{Mkt-Rf}, the market-cap weighted return of US equities minus the risk free rate,
\item \emph{SMB}, the return of a portfolio that is long small stocks and short big stocks,
\item \emph{HML}, the return of a portfolio that is long
value stocks and short growth stocks, and
\item \emph{Momentum}, the return of a portfolio that is long
high momentum stocks and short low (or negative) momentum stocks.
\end{itemize}
The daily returns have small enough means that they can be ignored.

\begin{figure}
\centering
\includegraphics[width=.7\textwidth]{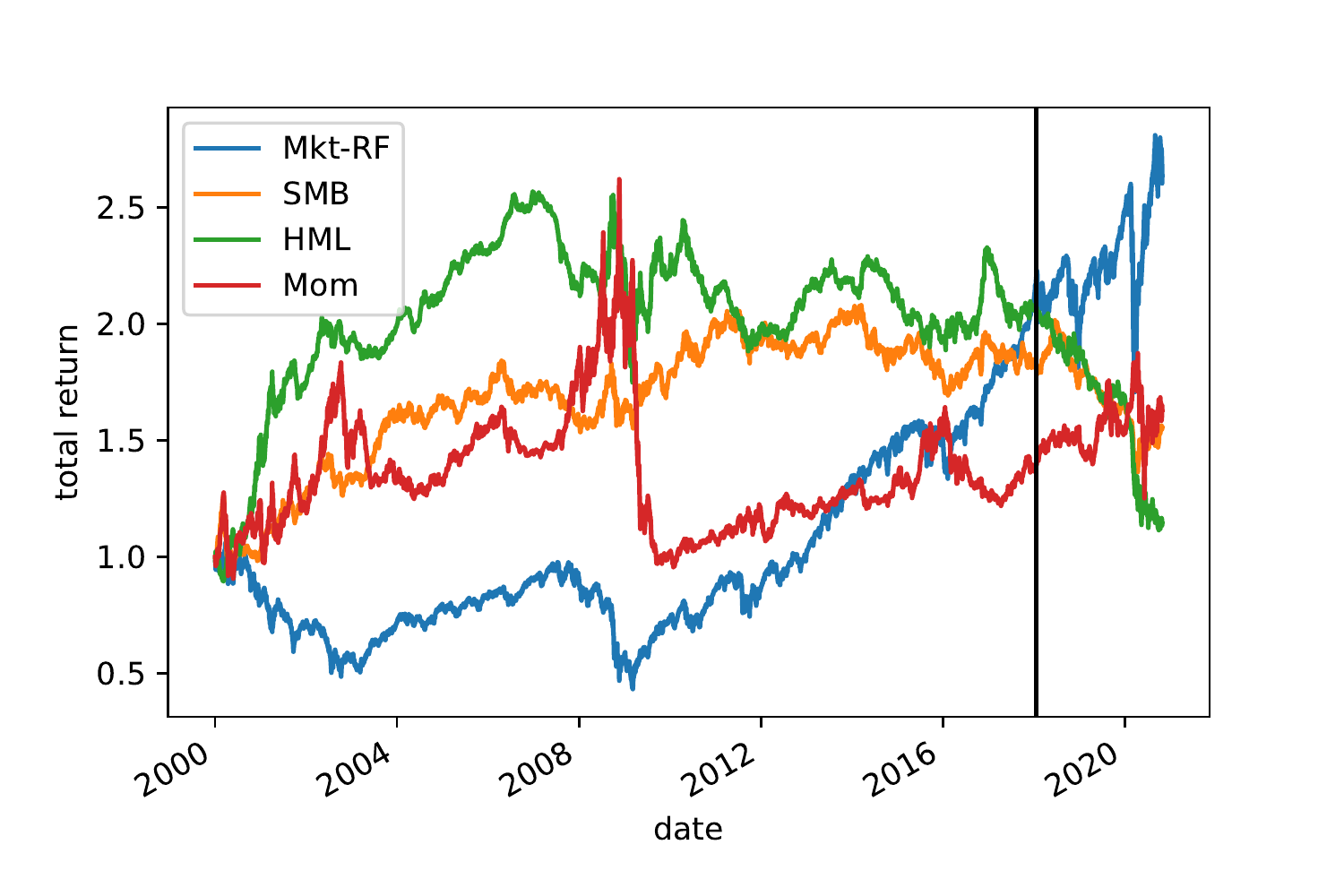}
\caption{The cumulative return of the four factors from 2000--2020.
The vertical black line denotes the split
between the train and test samples.}
\label{fig:factor_return}
\end{figure}

Our dataset runs from 2000--2020.
We split the dataset into a training dataset from 2000--2018 (4541 samples)
and a test dataset from 2018--2020 (700 samples).
The cumulative return of these four factors
(\ie, $\prod_{\tau=1}^t(1+(y_\tau)_i)$) from 2000--2020
is shown in figure \ref{fig:factor_return}.

\begin{figure}
\centering
\includegraphics[width=.7\textwidth]{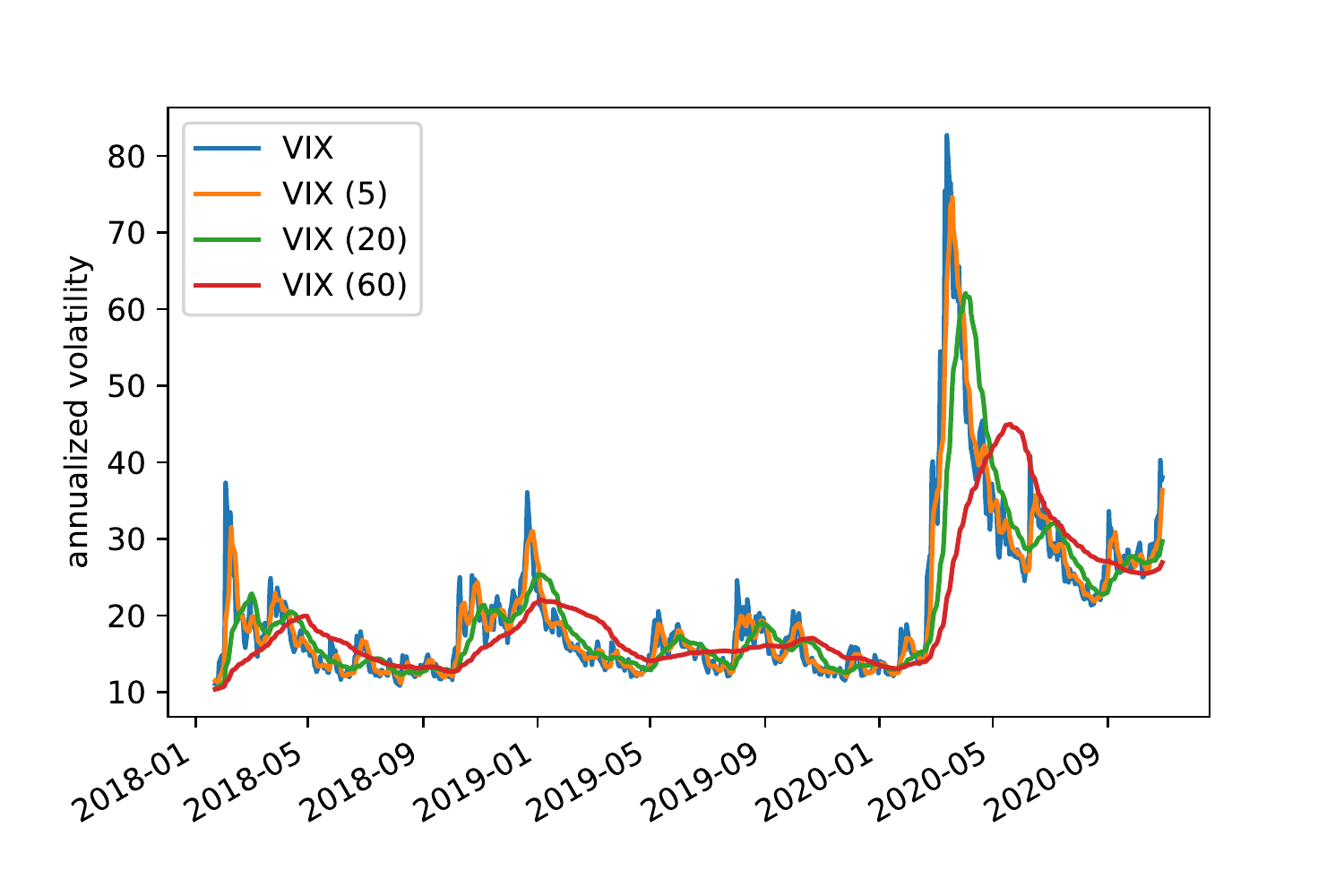}
\caption{The four VIX features over the test set.}
\label{fig:vix_features}
\end{figure}

\paragraph{VIX features.}
Our covariance models use several features derived 
from the CBOE volatility index (VIX),
a market-derived measure of expected 30-day
volatility in the US stock market.
We use the previous close of VIX as our raw feature.
We perform a quantile transform of VIX based on the training dataset, 
mapping it into $[-1,1]$ as described
in \S\ref{s-inverse_cholesky_model}. 
We will also use 5, 20, and 60 day trailing averages of VIX (which 
correspond to one week, around one month, and around one quarter).
These features are also quantile transformed to $[-1,1]$.
These four features are shown, before quantile transformation,
in figure \ref{fig:vix_features}. 

\begin{figure}
\centering
\includegraphics[width=.7\textwidth]{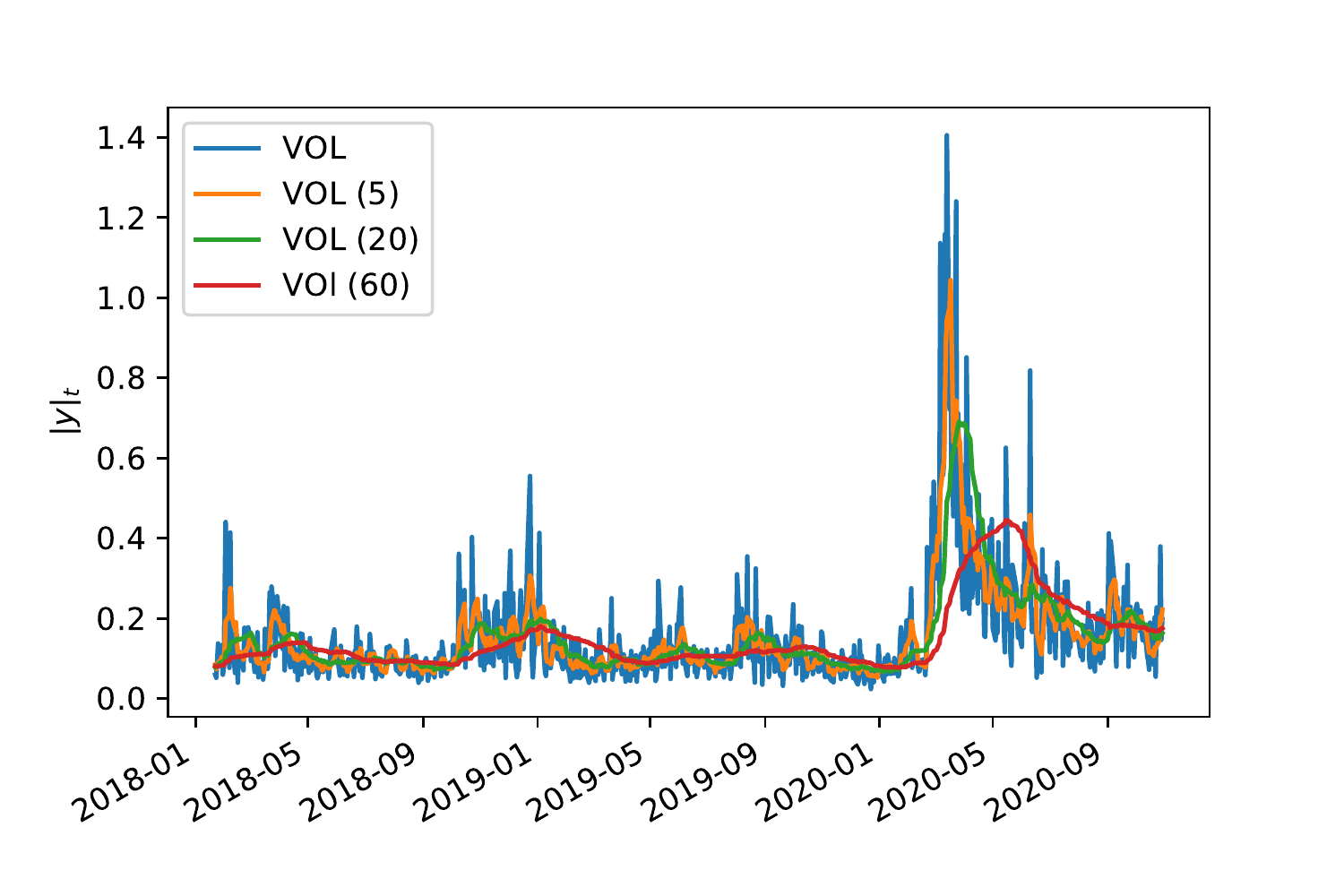}
\caption{The four VOL features over the test set.}
\label{fig:vix_rlz_features}
\end{figure}

\paragraph{VOL features.}
We use several features derived from previous realized returns,
which measure volatility.
One is the
sum of the absolute daily returns of the four factors over the 
previous day, \ie, $\|y_{i-1}\|_1$.
We also use trailing 5, 20, and 60 day averages of this quantity.
These four features are each quantile transformed and mapped 
into $[-1,1]$.
These four features are shown in figure \ref{fig:vix_rlz_features}, 
before quantile transformation.

\subsection{Covariance predictors}
We experiment with seven covariance prediction methods, organized into
three groups.

\paragraph{Simple predictors.}
\begin{itemize}
 \item \emph{Constant.} Fit a single covariance matrix to the training set.
 \item \emph{SMA.} We use memory $M=50$, 
which achieved the highest log-likelihood on the training set.
\end{itemize}

\paragraph{Regression whitener predictors.}
These predictors are based off the whitener regression approach described in \S\ref{s-inverse_cholesky_model}; we use the regularization function
\[
\lambda_1 (\|A\|_F^2 + \|C\|_F^2) + \lambda_2 (\|b - \ones\|_2^2 + \|d\|_2^2),
\]
for $\lambda_1,\lambda_2 > 0$.
The hyper-parameters $\lambda_1,\lambda_2$ are selected via a coarse grid search.
In all cases, we use $\epsilon=10^{-6}$.
\begin{itemize}
 \item \emph{VIX.} A regression whitener predictor with one feature, 
VIX. 
We use $\lambda_1=\lambda_2=0$.
 \item \emph{TR-VIX.} A regression whitener predictor with four features: 
VIX, and 5/20/60-day trailing averages of VIX.
We use $\lambda_1=10^{-5}$ and $\lambda_2=0$.
 \item \emph{TR-VIX-VOL.} A whitener regression predictor with eight features: 
VIX and 5/20/60-day trailing averages of VIX, and also $\|y_{i-1}\|_1$, 
and 5/20/60 day trailing averages.
We use $\lambda_1=10^{-5}$ and $\lambda_2=0$.
\end{itemize}

\paragraph{Iterated predictors.}
\begin{itemize}
 \item \emph{SMA, then TR-VIX-VOL.}  We first whiten with SMA with memory 50,
then with regression using TR-VIX-VOL.
For the regression predictor, we use $\lambda_1=10^{-5}$ and $\lambda_2=10^4$.
 \item \emph{TR-VIX-VOL, then SMA.} We first whiten with a regression with TR-VIX-VOL,
then with SMA, with memory 50.
For the regression predictor, we use $\lambda_1=10^{-5}$ and $\lambda_2=0$.
\end{itemize}

\subsection{Results}
The train and test log-likelihood of the seven covariance predictors
are reported in table \ref{tab:summary}.
We can see that a simple moving average with memory 50 does well,
in fact, better than the basic predictors based on whitener regressions
of VIX and features derived from VIX.
However, the iterated whitening predictors, SMA followed by TR-VIX,
does somewhat better,
with TR-VIX followed by SMA
doing the best.
This predictor gives an increase in likelihood over the SMA predictor of 
$\exp(14.1 - 13.59) = 1.67$, \ie, a 67\% lift.

\begin{table}
  \caption{Performance of seven covariance predictors on train and 
test sets.}
  \centering
  \begin{tabular}{lcc}
    \toprule
Predictor & Train log-likelihood & Test log-likelihood \\
\midrule
Constant & 13.60 & 12.18 \\
SMA (50) & 14.81 & 13.59 \\
VIX & 14.37 & 13.23 \\
TR-VIX & 14.40 & 13.32 \\
TR-VIX-VOL & 14.64 & 13.48 \\
SMA, then TR-VIX-VOL & 14.87 & 13.78 \\
TR-VIX-VOL, then SMA & 15.03 & 14.10 \\
\bottomrule
  \end{tabular}
  \label{tab:summary}
\end{table}

\begin{figure}
\centering
\includegraphics[width=\textwidth]{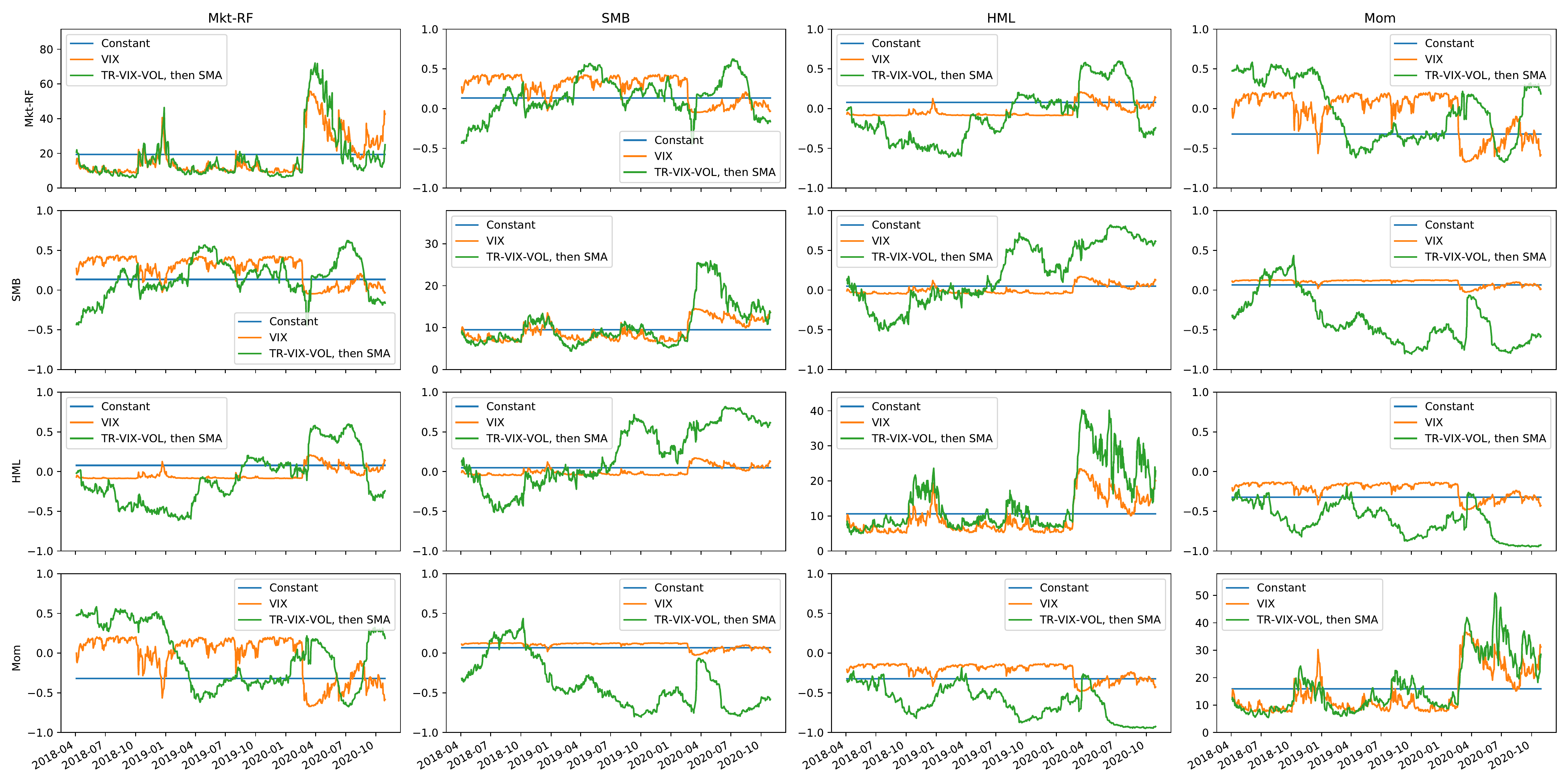}
\caption{Predicted annualized volatilities (on the diagonal) 
and correlations (on the off-diagonal) 
of the four factors from some of the methods.}
\label{fig:factors}
\end{figure}

\paragraph{Predicted covariances.}
Figure~\ref{fig:factors} shows the predicted volatilities and correlations
of three of the covariance predictors over the test set, with the 
volatilities given in annualized percent, 
\ie, $100 \sqrt{250 \Sigma_{ii}}$.
(The number of trading days in one year is around 250.)
The ones that achieve high test log-likelihood
vary considerably, with several correlations changing sign
over the test period.

\paragraph{Effect of ordering outcome components.}
Five of the predictors used in this example include the 
regression whitener, which as mentioned above depends on the 
ordering of the components in $y$.   
In each case, we tried all $4! = 24$ permutations (using the 
permutation whitener class), and found only negligible differences among them.
For all 24 permutations, 
the TR-VIX-VOL, then SMA predictor, achieved the top test performance 
log-likelihood,
with test log-likelihood ranging from 14.072 to 14.105.

\begin{figure}
\centering
\includegraphics[width=\textwidth]{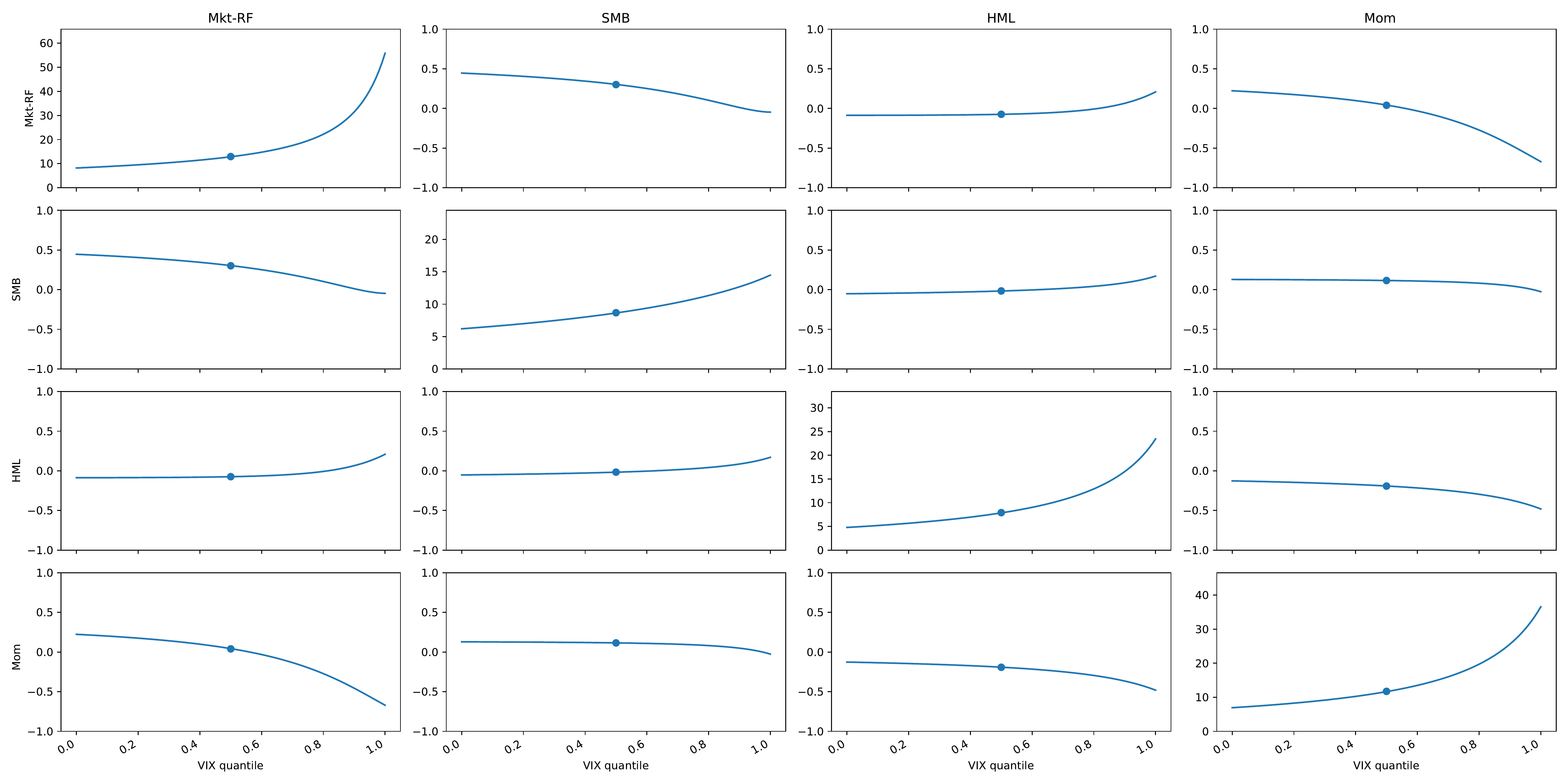}
\caption{Predicted annualized volatilities (on the diagonal) 
and correlations (off-diagonal) of the four factors versus VIX quantile,
for the VIX regression covariance predictor.
The dot represents the volatility or correlation when VIX is at its median.}
\label{fig:vix_forecast}
\end{figure}

\paragraph{The simple VIX regression predictor.}
The simple VIX regression model is readily interpretable.
Our predictor is
\[
\begin{bmatrix}
  129.4 & 0 & 0 & 0\\
  -58.3 & 184.2 & 0 & 0\\
  14.8 & -1.0 & 205.1 & 0\\
  0.8 & -15.9 & 26.4 & 135.7
\end{bmatrix}
+ x
\begin{bmatrix}
  -90.5 & 0 & 0 & 0\\
  64.6 & -73.2 & 0 & 0\\
  -1.8 & -13.1 & -128.2 & 0\\
  43.1 & 12.7 & -2.7 & -92.5
\end{bmatrix},
\]
where $x\in[-1,1]$ is the (transformed) quantile of VIX.
The lefthand matrix is the whitener when $x=0$, \ie, VIX 
takes its median value.
The righthand matrix shows how the whitener changes with $x$.
For example, as $x$ varies over its range $[-1,1]$,
$(L)_{11}$ varies over the range $[38.9,220.0]$, a 
factor of around of 5.7.
We can easily understand how the predicted covariance changes
as $x$ (the quantilized shifted VIX) varies.
Figure \ref{fig:vix_forecast} shows the predicted volatilities
(on the diagonal) and correlations (on the off-diagonals)
as the VIX feature ranges over $[-1,1]$.
We see that as VIX increases, all the predicted volatilities increase.
But we can also see that VIX has an effect on the correlations.
For example, when VIX is low, the momentum factor is positively 
correlated with the market factor;
when VIX is high, the momentum factor becomes negatively correlated 
with the market factor, according to this model.

\subsection{Multi-day covariance predictions}\label{s-multi-day}
The covariance predictors above predict the covariance of the 
return over the next trading day, \ie, $y_i$.
In this section we form covariance predictors for the next 1, 
20, 60, and 250 trading days.
(The last three correspond to around one calendar month, quarter, and year.)
As mentioned in \S\ref{s-multiple-outcomes}, 
this is easily done with the same model, by replicating each data point.
For example, to predict a covariance matrix for the next 5 days,
we take the data record $(x_i, y_i)$ and form five data records,
\[
(x_i, y_i), ~(x_i,y_{i+1}), ~\ldots,~ (x_i, y_{i+4}),
\]
and then use our method to predict the covariance.

\begin{table}
  \caption{Train and test log-likelihood of the 1, 20, 60, and 250-day predictors.}
  \centering
  \begin{tabular}{lll}
    \toprule
Days & Train log-likelihood & Test log-likelihood \\
\midrule
1 & 14.36 & 13.29 \\
20 & 14.24 & 12.83 \\
60 & 14.09 & 11.93 \\
250 & 13.91 & 11.75 \\
\bottomrule
  \end{tabular}
  \label{tab:summary3}
\end{table}

We form multi-day covariance predictions over the next 1, 20, 60, and 250 training days for the VIX regression predictor.
We report the train and test log-likelihoods of each of these predictors
in table \ref{tab:summary3}.
As expected, the log-likelihood decreases as the number of days ahead we need to predict increases.
Figure \ref{fig:vix_forecast} shows the predicted volatilities
(on the diagonal) and correlations (on the off-diagonals)
as the VIX feature ranges over $[-1,1]$
for the 1, 20, 60, and 250-day predictors.
We see that the predicted volatility for the market factor over the 
next day is much more sensitive to VIX than the predicted volatility 
over the next 250 days.
This suggests that volatility is mean-reverting, \ie, 
over the long run, volatility tends to return to its mean value.
We also observe a similar phenomenon with
the correlations, although it is less pronounced;
for example, the correlation between the HML and momentum factor can go 
from $-0.2$ to $-0.5$ based on VIX on up to 60-day horizons, 
but stays more or less constant at $-0.3$ over the 250-day horizon.

\begin{figure}
\centering
\includegraphics[width=\textwidth]{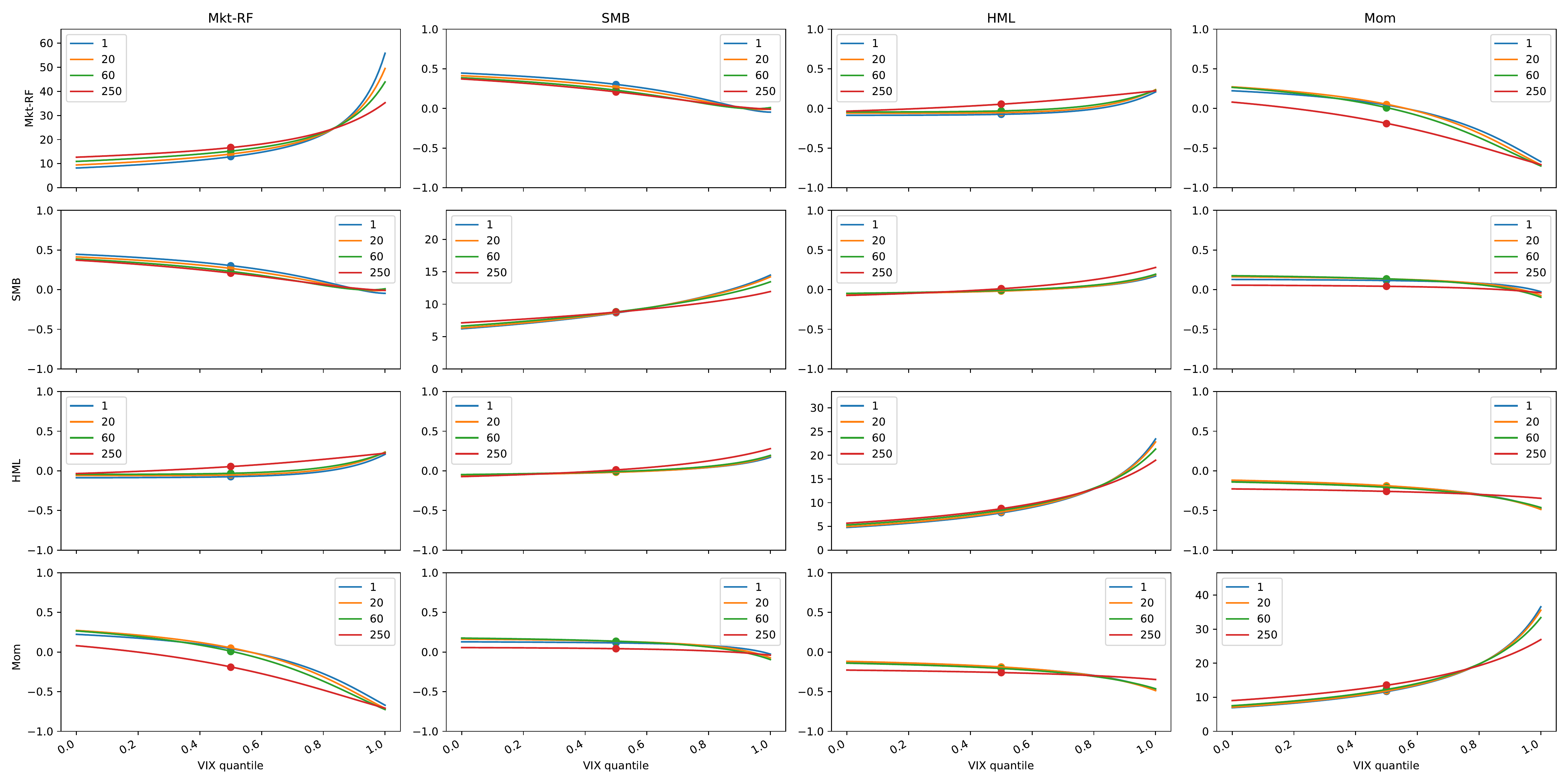}
\caption{Predicted annualized volatilities (on the diagonal) 
and correlations (off-diagonal) of the four factors versus VIX quantile,
for the VIX regression covariance predictor for four horizons: 1, 20, 60, and 250 days.
The dot represents the predicted volatility or correlation when VIX is at its median.}
\label{fig:vix_forecast_horizon}
\end{figure}

\clearpage
\section{Example: Machine learning residuals} \label{s-ml-resid}

In this section we present an example where the predicted covariance 
is of the prediction error or residuals of a point predictor.

\subsection{Outcome and features}

\paragraph{Dataset.}
We consider the ``Communities and Crime'' dataset from the UCI machine learning repository
\cite{ml_example_1,ml_example_2,ml_example_3,ml_example_4,ml_example_5}.
The dataset consists of 128 attributes of 1994 communities within the United States.
These attributes describe the demographics of the community, 
as well as the socio-economic conditions and crime statistics.
We removed the attributes that are categorical or have missing values, 
leaving 100 attributes.
All attributes came normalized in the range $[0,1]$.
We randomly split the dataset into a 1495-sample training dataset and 
a 499-sample test dataset.

\paragraph{Outcome.}
We choose the following two attributes to be the outcome:
\begin{itemize}
	\item \emph{agePct65up}. The fraction of the population age 65 and up.
	\item \emph{pctWSocSec}. The fraction of the population that has social security income.
\end{itemize}
(These were intentionally picked because they have non-trivial correlation.)
We map each of these two attributes to
have unit normal marginals
on the training set by quantilizing each feature and then applying the inverse CDF of the unit normal.

\paragraph{Features.}
We use the remaining 98 attributes as the features.
We use a quantile transformation for each of these
using the training set, and map
the resulting features in the train and test set to $[-1,1]$.

\subsection{Regression residual covariance predictors}
In our first example we take the simple approach mentioned in
\S\ref{s-nonzero-mean}, where we first form a predictor of the mean, and then form a model of the covariance of the residuals.

\paragraph{Ridge regression model.}
We fit a ridge regression model to predict the two output attributes from the 98 input attributes, using cross validation on the training set to select the regularization parameter.
The root mean squared error (RMSE) of this model on the training set was 0.352 and on the test set was 0.359.

\paragraph{Regression residuals.}
We use the residuals from the regression model as $y_i$.
That is, if $y^\mathrm{true}_i$ is the true outcome, and our regression model predicts $\hat y_i$, then we let $y_i = y^\mathrm{true}_i - \hat y_i$.
Our goal is to model the covariance of the residuals $y_i$ 
using $x_i$ as features.

We experiment with seven covariance prediction methods, organized into three groups.

\paragraph{Constant predictor.}
Fit a single covariance matrix to the training set. This covariance matrix was
 \[
\Sigma = \begin{bmatrix}
  0.14 & 0.08\\
  0.08 & 0.11\\
\end{bmatrix}.
 \]

\paragraph{Regression whitener predictors.}
Hyper-parameters were selected via a very coarse grid search.

\begin{itemize}
 \item \emph{Diagonal.} The diagonal predictor in \eqref{eq:diagon_model} with the regularization function $(0.1)\|A\|_F^2$.
 \item \emph{Regression.} The regression whitener predictor in \S\ref{s-inverse_cholesky_model} with $\epsilon=10^{-2}$ and the regularization function $(0.01)(\|A\|_F^2 + \|C\|_F^2)$.
\end{itemize}

\paragraph{Iterated predictors.}
Hyper-parameters were selected via a very coarse grid search.
\begin{itemize}
 \item \emph{Constant, then diagonal.} The constant predictor, followed by a diagonal predictor with the regularization function $(0.1)\|A\|_F^2$.
 \item \emph{Diagonal, then constant.} The diagonal predictor with the regularization function $(0.1)\|A\|_F^2$, followed by a constant predictor.
 \item \emph{Constant, then regression.} The constant predictor, followed by a whitener regression predictor with $\epsilon=10^{-2}$ and the regularization function $\|A\|_F^2 + \|C\|_F^2 + \|b-1\|_2^2 + \|d\|_2^2$.
 \item \emph{Regression, then constant.} The whitener regression predictor with $\epsilon=10^{-2}$ and the regularization function $(0.1)(\|A\|_F^2 + \|C\|_F^2)$, followed by a constant predictor.
\end{itemize}

\begin{table}
  \caption{Performance of seven covariance predictors on train and 
test sets.}
  \centering
  \begin{tabular}{lcc}
    \toprule
Predictor & Train log-likelihood & Test log-likelihood \\
\midrule
Constant & -0.47 & -0.44 \\
Diagonal & -0.37 & -0.55 \\
Regression & -0.05 & -0.14 \\
Constant, then diagonal & -0.45 & -0.69 \\
Diagonal, then constant & 0.00 & -0.18 \\
Constant, then regression & -0.26 & -0.33 \\
Regression, then constant & 0.01 & -0.11 \\
\bottomrule
  \end{tabular}
  \label{tab:summary5}
\end{table}

\subsection{Results}
The train and test log-likelihood of the seven covariance predictors
are reported in table \ref{tab:summary5}.
We can see that the diagonal predictor does the worst, 
likely because the diagonal predictor fails to model the substantial correlation 
between the outcomes.
The whitener regression predictor does much better than the constant predictor,
with a lift of 35\% in likelihood.
The best predictor was the regression whitener, then constant predictor, 
with a lift of 39\% in likelihood over the constant predictor.

\paragraph{Covariance variation.}
The regression, then constant
predictor predicts a different covariance matrix for each residual,
which varies significantly over the test dataset.
The standard deviation of the first component 
varies over the range $[0.19,1.16]$, and the 
standard deviation of the second component 
varies over the range $[0.18,0.88]$.
The correlation between the two components varies over the range 
$[-0.045, 0.908]$, \ie, from slightly negatively correlated to 
strongly correlated.

\begin{figure}
\centering
\includegraphics[width=.8\textwidth]{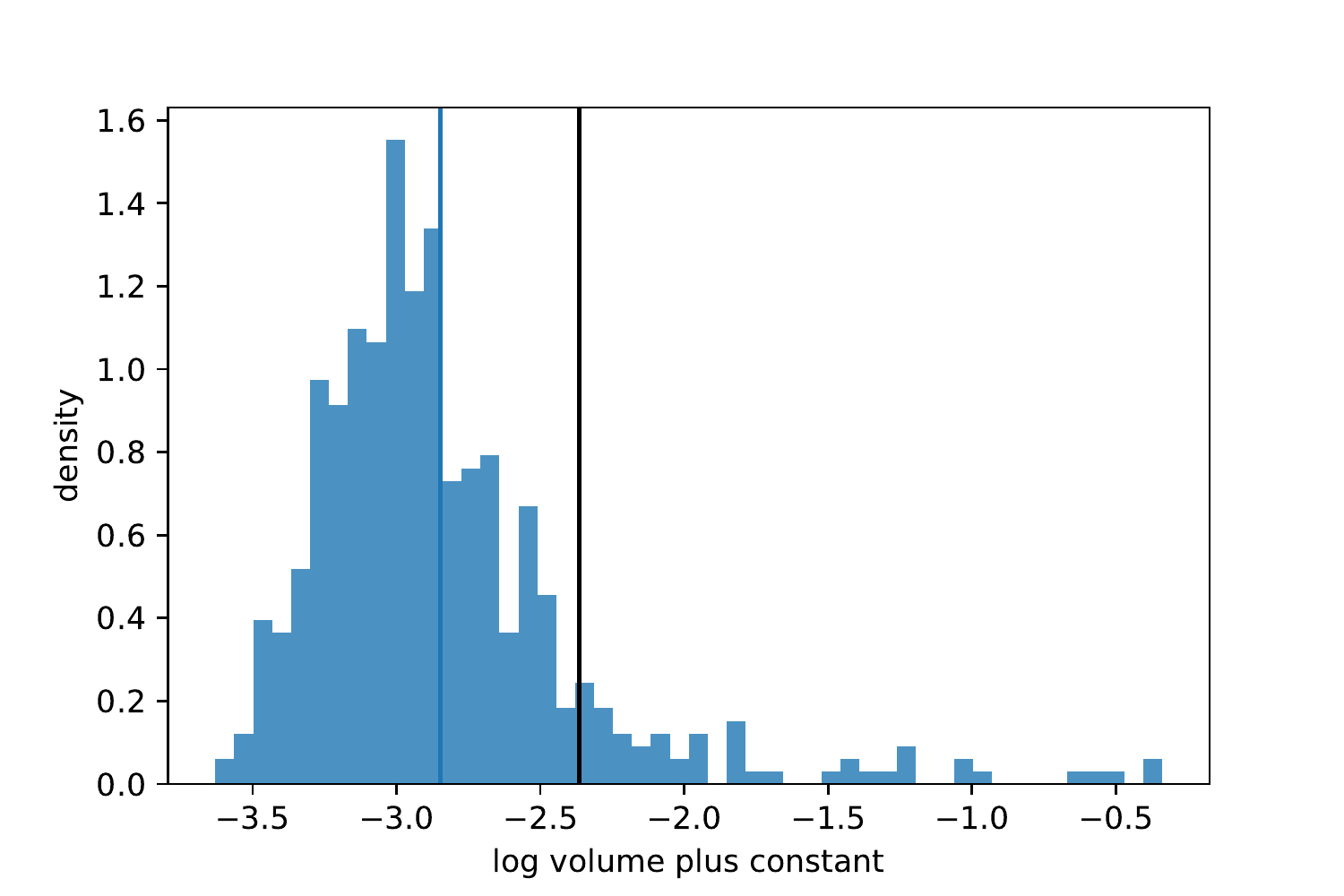}
\caption{Log volume of the regression, then constant predictor over the test set. The blue vertical line is the average log volume, and the black vertical line is the log volume of the constant predictor.}
\label{fig:log_volume}
\end{figure}

\paragraph{Volume of the confidence ellipsoids.}
The confidence regions of a multivariate Gaussian are ellipsoids, with 
volume proportional to $\det(\Sigma)^{1/2}$.
In figure \ref{fig:log_volume} we plot the log volume (which here is area since
$n=2$) plus a constant of the 
`regression, then constant' predictor over the test set, 
along with the equivalent log volume of the constant predictor.
Most of the time, the volume of the confidence ellipsoid predicted by 
the best predictor is smaller than the constant predictor.
Indeed, on average, the confidence ellipsoid occupies 62\% of the area.

\begin{figure}
\centering
\includegraphics[width=.8\textwidth]{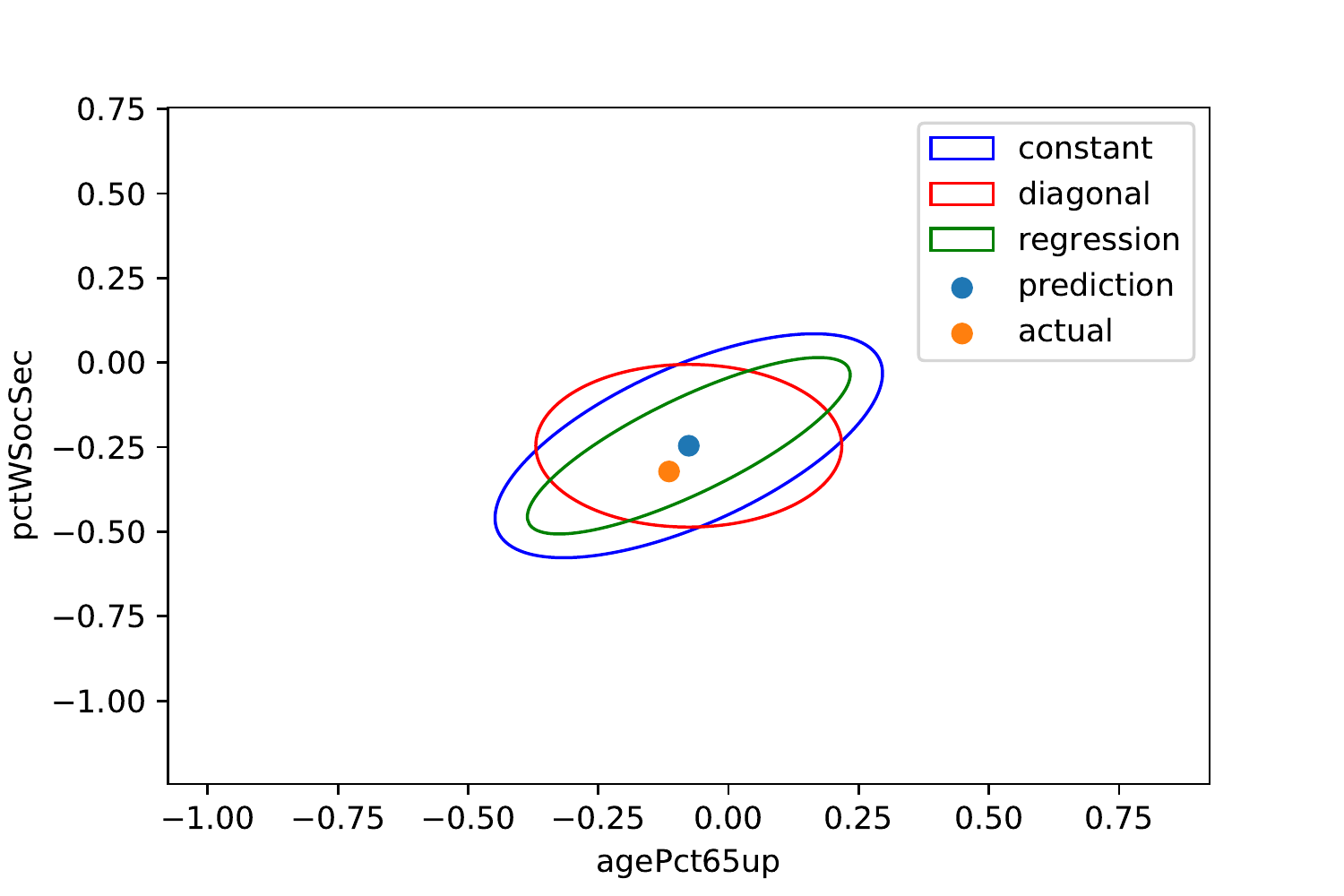}
\caption{Confidence ellipsoid of a test sample for three covariance predictors, and the actual outcome.}
\label{fig:confidence_ellipse}
\end{figure}

\begin{figure}
\centering
\includegraphics[width=.8\textwidth]{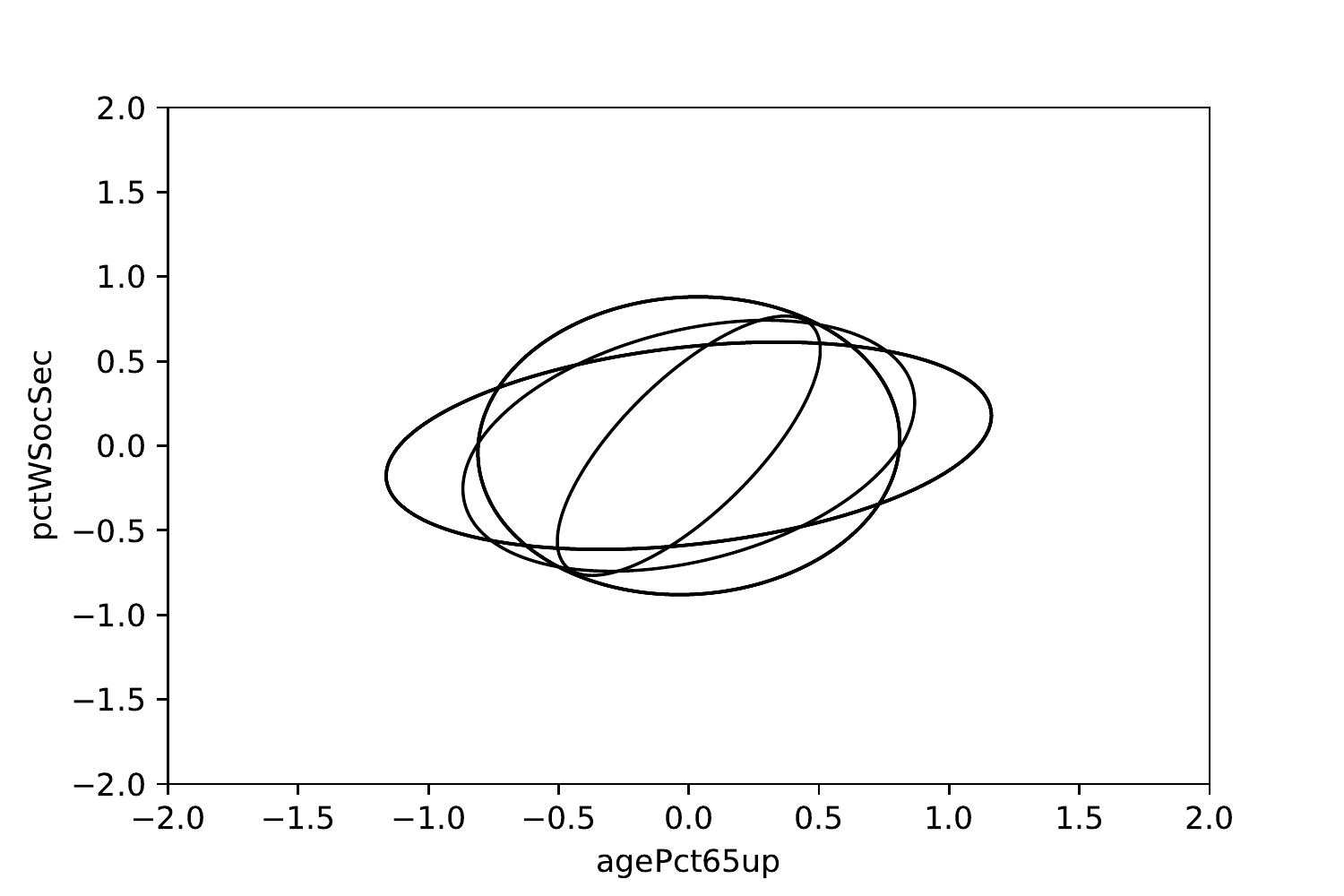}
\caption{Some extreme confidence ellipsoids on the test set for the `regression, then constant' predictor.}
\label{fig:confidence_ellipse_extremes}
\end{figure}

\paragraph{Visualization of confidence ellipsoids.}
In figure \ref{fig:confidence_ellipse} we plot the one-$\sigma$
confidence ellipsoids of predictions, on a particular sample from the test dataset.
We see that the area of the constant ellipsoid is much larger than the two other predictors,
and that the diagonal predictor does not predict any correlation between the outcomes.
However, the regression predictor does predict a correlation, and significantly less standard deviation in both outcomes.
For this particular example, the regression predictor confidence region is 51\%
the area of the constant predictor.
In figure \ref{fig:confidence_ellipse_extremes} we visualize some 
(extreme) one-$\sigma$ confidence ellipsoids of the `regression, then constant' 
predictor on the test set, demonstrating how much they vary.

\clearpage
\subsection{Joint mean-covariance prediction}
In this section we perform joint prediction of the conditional mean and 
covariance as described in \S\ref{s-nonzero-mean}.
We solve the convex problem
\BEQ
\begin{array}{ll}
\text{maximize} & (1/N)\sum_{i=1}^{N}\left( \sum_{j=1}^n \log (L_i)_{jj} - (1/2)
\|L_i^Ty_i-\nu_i\|_2^2 \right) - (0.1)(\|A\|_F^2 + \|C\|_F^2)\\
\text{subject to} & \mathbf{diag}(L_i) = Ax_i + b, \quad i=1,\ldots,N,\\
& \mathbf{offdiag}(L_i) = Cx_i + d, \quad i=1,\ldots,N,\\
& \nu_i = Ex_i + f, \quad i=1,\ldots,N,\\
& \|A\|_{\text{row},1}\leq b - \epsilon,
\end{array}
\label{eq:prob_joint}
\EEQ
with variables $(A,b,C,d,E,f)$ using L-BFGS-B.

By jointly predicting the conditional mean and covariance, 
we actually achieve a better test RMSE than predicting just the mean.
The train RMSE of this model was 0.273 and the test MSE
was 0.331, whereas the
RMSE of the ridge regression model was 0.352 on the training set 
and 0.359 on the test set.
Thus, jointly modeling the mean and covariance results in a 7.8\% reduction in 
RMSE on the test set.

\begin{figure}
\centering
\includegraphics[width=.8\textwidth]{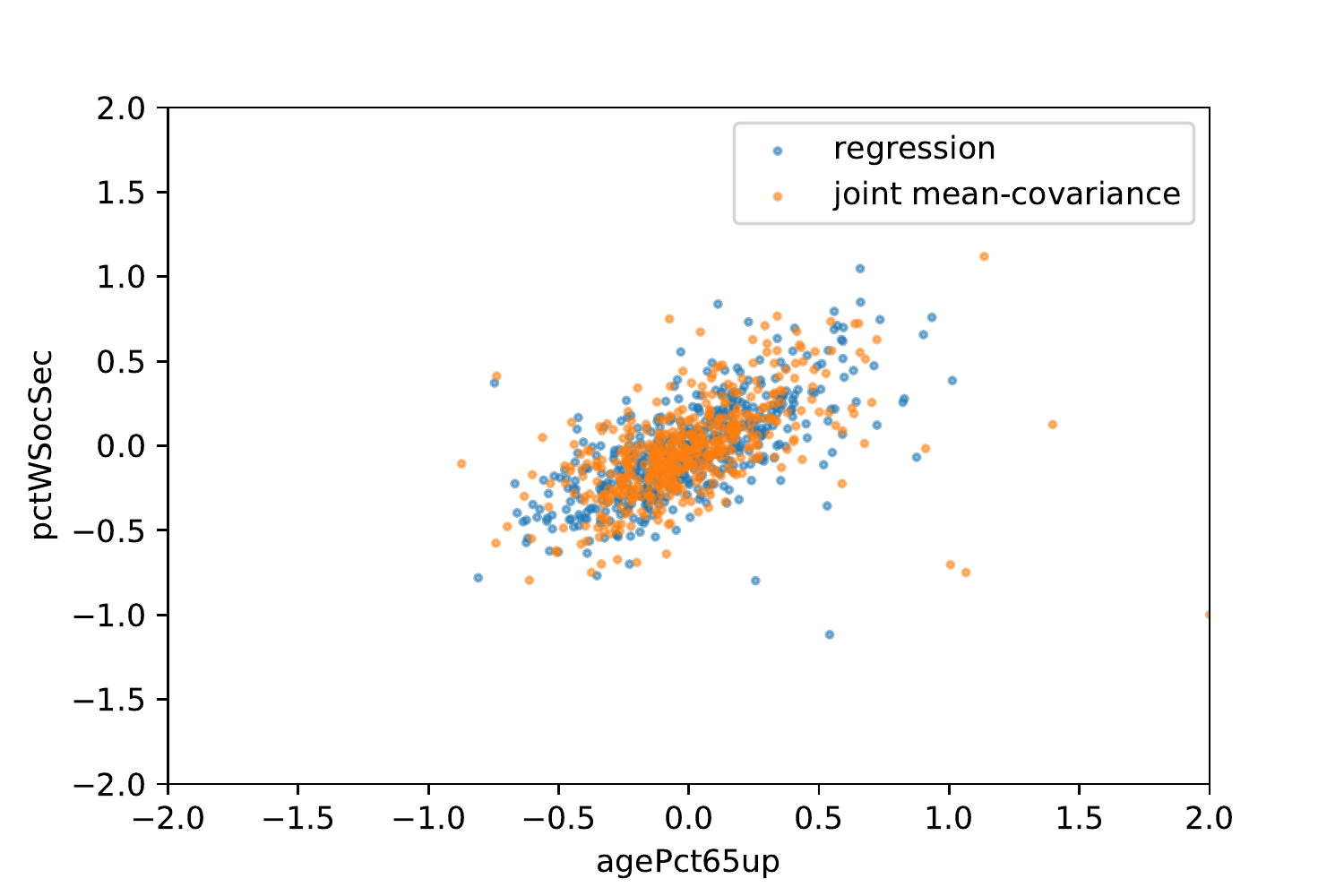}
\caption{Test set residuals for the regression predictor and joint mean-covariance predictor.}
\label{fig:joint_residuals}
\end{figure}

In figure \ref{fig:joint_residuals} we visualize the residuals of the ridge regression model and the joint mean-covariance model on the test set.
We observe that the joint mean-covariance residuals seem to be on average closer to the origin.
In terms of Gaussian log-likelihood, the ridge regression model with a constant covariance achieves a train log-likelihood of 0.054 and a test log-likelihood of 0.088.
In contrast, the joint mean-covariance model achieves a train log-likelihood of 1.132 and a test log-likelihood of 1.049, representing a lift on the test set of 161\%.

\begin{figure}
\centering
\includegraphics[width=.8\textwidth]{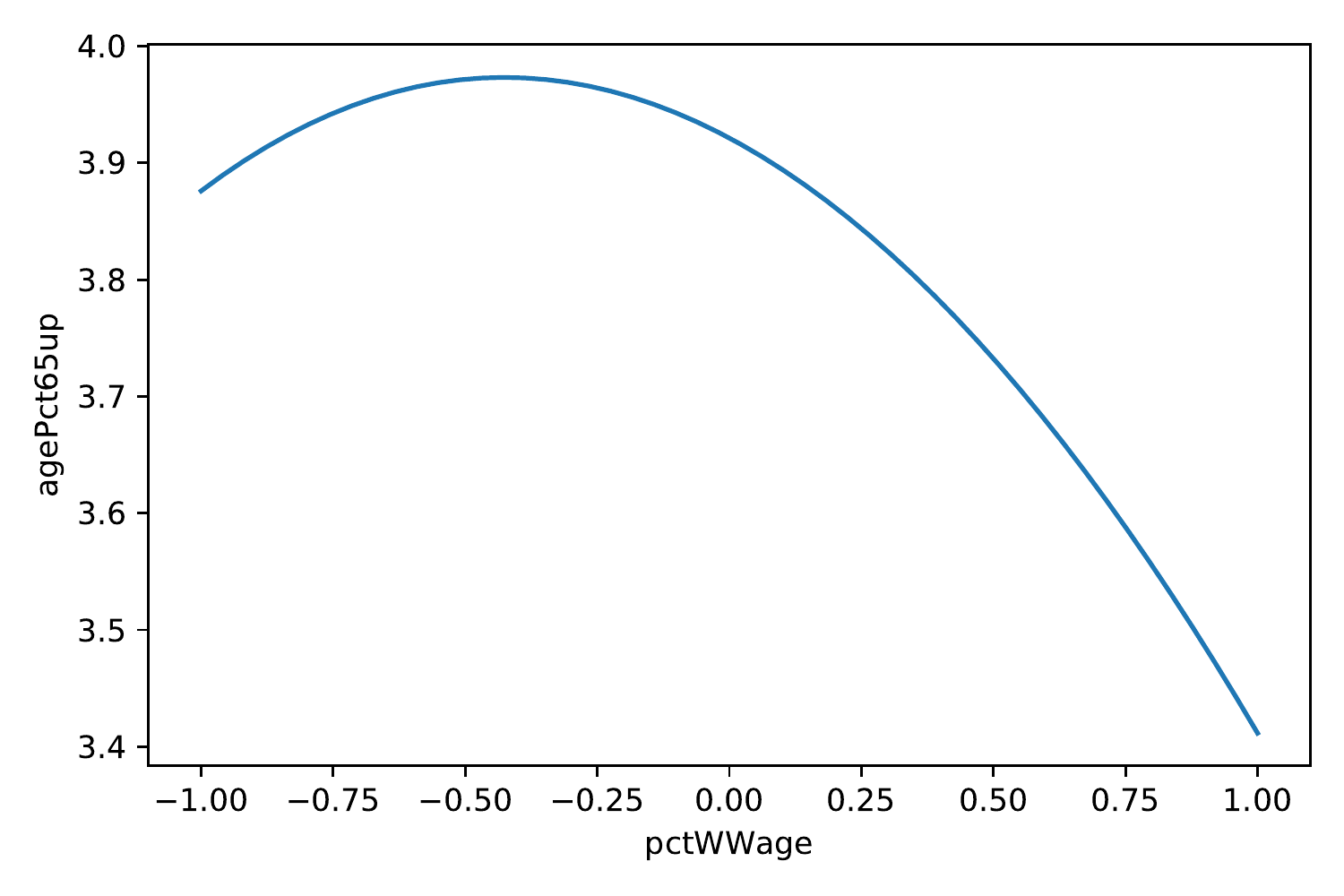}
\caption{The nonlinear effect of a particular feature on the mean prediction in the joint mean-covariance model.}
\label{fig:feature_effect}
\end{figure}

Recall that the prediction of the mean by the joint mean-covariance model 
is nonlinear in the input.
In figure \ref{fig:feature_effect} we visualize this effect
for a particular test point, by varying just the `pctWWage' feature from $-1$ to $1$,
and visualizing the change in the mean prediction for the `agePct65up' output.
The nonlinearity is evident.

\clearpage
\section{Conclusions and future work}

Many covariance predictors, ranging from simple to complex,
have been developed.  Our focus has been on the regression 
whitener, which has a concave log-likelihood function,
so fitting reduces to a convex optimization problem that is 
readily solved.
The regression whitener is also readily interpretable,
especially when the number of features is small, or a 
rank-reducing regularizer results in a low rank coefficient matrix
in the predictor.
Among other predictors that have been proposed, the only other ones
that share the property of having a concave log-likelihood is
the diagonal (exponentiated) covariance predictor and
the Laplacian regularized stratified predictor.

We observed that covariance predictors can be iterated;
our examples show that simple sequences of predictors can indeed
yield improved performance.
While iterated covariance prediction can yield
better covariance predictors, it raises the question of how to choose
the sequence of predictors.  At this time, we do not know, and 
can only suggest a trial and error approach.
We can hope that this question is at least partially answered
by future research.

As other authors have observed, it would be nice to identify
an unconstrained parametrization of covariance matrices, \ie,
an inverse link mapping that maps all of $\reals^p$ onto $\symm_{++}^n$.
(Our regression whitener requires the constraint $\|x\|_\infty \leq 1$.)
One candidate is the matrix exponential, which maps symmetric
matrices onto $\symm_{++}^n$.  Unfortunately, this parametrization
results in a log-likelihood that is not concave.
As far as we know, the existence of an 
unconstrained parametrization
of covariance matrices, with a concave log-likelihood, 
is still an open question.

Finally, we mention one more issue that we hope will be 
addressed in future research.
The dependence of the regression whitener 
on the ordering of the entries of $y$ 
certainly detracts from its aesthetic and theoretical appeal.
In our examples, however, we have obtained similar results
with different orderings, suggesting that the dependence on
ordering is not a large problem in practice, though it should still be checked.
Still, some guidelines as to how to choose the ordering,
or otherwise address this issue, would be welcome.

\section*{Acknowledgements}

The authors gratefully acknowledge conversations and discussions about 
some of the material in this paper with Misha van Beek, Linxi Chen, David Greenberg,
Ron Kahn, Trevor Hastie, Rob Tibshirani, Emmanuel Candes, Mykel Kochenderfer, and 
Jonathan Tuck.

\clearpage
\bibliography{citations}

\end{document}